\documentclass[10pt,twocolumn,letterpaper]{article}

\usepackage{cvpr}              

\usepackage{amsmath,amssymb}
\usepackage{graphicx}
\usepackage[ruled,vlined,linesnumbered]{algorithm2e} 
\usepackage{booktabs}
\usepackage{siunitx}
\usepackage{xspace}
\usepackage[table]{xcolor}
\newcommand{\SectionRow}[1]{%
  \rowcolor{gray!12}%
  \multicolumn{22}{l}{\textit{#1}}\\[1pt]
}
\usepackage{multirow}
\usepackage{pgfplots}
\pgfplotsset{compat=1.18}
\definecolor{oiBlue}{HTML}{0072B2}
\definecolor{oiSky}{HTML}{56B4E9}
\definecolor{oiOrange}{HTML}{E69F00}
\definecolor{oiGreen}{HTML}{009E73}
\definecolor{oiYellow}{HTML}{F0E442}
\definecolor{oiVermilion}{HTML}{D55E00}
\definecolor{oiPurple}{HTML}{CC79A7}
\pgfplotsset{
  cycle list/.define={
    {oiBlue, mark=*},
    {oiOrange, mark=square*},
    {oiSky, mark=triangle*},
    {oiGreen, mark=diamond*},
    {oiVermilion, mark=otimes*},
    {oiPurple, mark=pentagon*},
    {oiYellow, mark=star}
  },
}
\usepackage{adjustbox}
\usepackage{booktabs}
\usepackage{xcolor}
\usepackage{pifont}
\usepackage{booktabs,xcolor,pifont}
\usepackage{tcolorbox}
\usepackage{inconsolata}

\newcommand{\cmark}{\textcolor{green!50!black}{\checkmark}}
\newcommand{\xmark}{\textcolor{red!70!black}{\ding{55}}}
\newcommand{\method}{ZoomClick\xspace} 

\definecolor{cvprblue}{rgb}{0.21,0.49,0.74}
\usepackage[pagebackref,breaklinks,colorlinks,allcolors=cvprblue]{hyperref}

\NewDocumentCommand{\shilong}
{ mO{} }{\textcolor{red}{\textsuperscript{\textit{Shilong}}\textsf{\textbf{\small[#1]}}}}

\def\paperID{1621} 
\def\confName{CVPR}
\def\confYear{2026}

\title{Zoom in, Click out: Unlocking and Evaluating the \\Potential of Zooming for GUI Grounding}

\author{
Zhiyuan Jiang\textsuperscript{1,2,3}\thanks{Equal contributions.}\hspace{0.3em}\thanks{Work done during an internship at Princeton AI$^2$ Lab \& PKU.}\hspace{0.4em}
Shenghao Xie\textsuperscript{3}\footnotemark[1]\hspace{0.3em}
Wenyi Li\textsuperscript{4}\hspace{0.3em}
Wenqiang Zu\textsuperscript{4,3}\hspace{0.3em}
Peihang Li\textsuperscript{5}\hspace{0.3em}
Jiahao Qiu\textsuperscript{2}\hspace{0.3em}  \\
Siqi Pei\textsuperscript{6}\hspace{0.3em}
Lei Ma\textsuperscript{3}\thanks{Corresponding authors. Contact: \texttt{sl8264@princeton.edu}}\hspace{0.5em}
Tiejun Huang\textsuperscript{3} \hspace{0.1em}
Mengdi Wang\textsuperscript{2}\footnotemark[3]\hspace{0.5em}
Shilong Liu\textsuperscript{2}\footnotemark[3]\\[0.5em]
\textsuperscript{1}Xi'an Jiaotong University \hspace{0.3em}
\textsuperscript{2}Princeton University \hspace{0.3em}
\textsuperscript{3}Peking University \\
\textsuperscript{4}University of Chinese Academy of Sciences \hspace{0.3em}
\textsuperscript{5}The University of Hong Kong \hspace{0.3em} \\
\textsuperscript{6}Michigan State University \\[1.4em]
}

\begin{document}
\maketitle

\begin{abstract}
    Grounding is a fundamental capability for building graphical user interface (GUI) agents. Although existing approaches rely on large-scale bounding box supervision, they still face various challenges, such as cross-platform generalization, complex layout analysis, and fine-grained element localization. In this paper, we investigate \textbf{zoom} as a strong yet underexplored prior for GUI grounding, and propose a training-free method, \textbf{ZoomClick}. By characterizing four key properties of zoom (\ie, pre-zoom, depth, shrink size, minimal crop size), we unlock its full capabilities for dynamic spatial focusing and adaptive context switching. Experiments demonstrate that our method significantly boosts the performance of both general vision–language and specialized GUI grounding models, achieving state-of-the-art results on several mainstream benchmarks, \eg, UI-Venus-72B attains a 73.1\% success rate on ScreenSpot-Pro. Furthermore, we present \textbf{GUIZoom-Bench}, a benchmark for evaluating model's adaptability to zoom, aiming to inspire future research on improving zoom for further training and test-time scaling in GUI grounding tasks. Code is available at \href{https://github.com/Princeton-AI2-Lab/ZoomClick}{https://github.com/Princeton-AI2-Lab/ZoomClick}.
\end{abstract}    
\vspace{-6mm}
\section{Introduction}

As digital platforms (\eg, desktops, webpages, and applications) increasingly serve as the primary medium for human-computer interaction (HCI), there has been growing interest in developing GUI agents \cite{liu2025paluiplanningactivelookback}\cite{lu2024omniparserpurevisionbased}\cite{openai2024gpt4technicalreport}\cite{xie2025scalingcomputerusegroundinguser, gao2025surveyselfevolvingagentspath, qiu2025alitageneralistagentenabling}, which understands user instructions, perceives screenshots, plans operation steps, and performs interactive actions. A fundamental capability of GUI agent perception is grounding \cite{cheng2024seeclickharnessingguigrounding}\cite{li2025screenspotproguigroundingprofessional}\cite{wu2024osatlasfoundationactionmodel, liu2024groundingdino, ren2024groundedsam, zhang2024llavagrounding, zhang2024fmov3d, ren2024groundingdino15advance}, mapping the textual descriptions of a target visual element to its on-screen coordinates, thereby generating bounding boxes for localization.

Existing approaches mainly leverage general multimodal large language models (MLLMs) \cite{bai2025qwen25vltechnicalreport}\cite{wang2024qwen2vlenhancingvisionlanguagemodels}\cite{yang2025qwen3technicalreport} or visual grounding models to finetune specialized GUI grounding models \cite{gu2025uivenustechnicalreportbuilding}\cite{qin2025uitarspioneeringautomatedgui}, and further introduce several post-training \cite{zhao2025learningguigroundingspatial, li2025eagle2buildingposttraining} or test-time scaling \cite{luo2025visualtesttimescalinggui}\cite{xu2024attentiondrivenguigroundingleveraging} strategies to enhance their performance\cite{ye2025guiarpenhancinggroundingadaptive}. Among these strategies, \textbf{zoom} \cite{nguyen2025improvedguigroundingiterative, qi2025cogcomvisuallanguagemodel} is intuitively appealing, since grounding requires not only understanding the global context\cite{nayak2025uivisiondesktopcentricguibenchmark} but also distinguishing fine-grained local regions\cite{li2025screenspotproguigroundingprofessional}, and zoom enables the model to dynamically switch between them.

However, the implementation of zoom in existing studies remains limited. For example, they either broadly aggregate predictions from a single zoom applied to multiple ROIs \cite{luo2025visualtesttimescalinggui}\cite{wu2025dimoguiadvancingtesttimescaling}, or recursively zoom into a specific region without an effective stopping criterion, which may result in missed small icons or produce excessively high-resolution crops that are out-of-distribution (OOD). Although certain studies optimize zoom via training, such as introducing IoU-aware losses to refine zoom ranges\cite{park2025rvlmregionawarevisionlanguage} or maintaining global-local consistency via zoom\cite{lee2025reguidedataefficientgui}, these approaches are still auxiliary techniques, as the intrinsic test-time scaling\cite{byun2025testtimescalingzeroshotdiagnosisvisuallanguage}\cite{kaya2025efficienttesttimescalingsmall} potential of zoom itself is largely underexplored.

In this paper, we explore several highly practical insights by conducting a comprehensive empirical analysis of zoom.

\textit{Insight 1: Zoom is simple yet effective, its full capability better activates the vision–language and grounding priors of foundation models.} We propose \textbf{ZoomClick}, a training-free method that integrates four critical properties of zoom (\ie, pre-zoom, depth, shrink size, and minimal crop size). Specifically, ZoomClick undergoes a three-stage pipeline. 

\textbf{When to zoom in.} ZoomClick begins with a Pre-Zoom that compares a global prediction with four local predictions over a $2\times2$ grid. A local candidate becomes the starting point when its distance to the global candidate falls below a threshold, ensuring early correction with full context still available and steering the model toward regions that receive high confidence both globally and locally.

\textbf{How to zoom.} Each iteration crops a region of the specified shrink size directly in the original coordinate system, preventing iterative drift from relative cropping and avoiding boundary overflow that introduces irrelevant context.

\textbf{When to click out.} Termination occurs when no target is detected, the zoom depth is reached, or the crop hits the minimal size, enabling resolution-adaptive stopping and preventing over-zooming into unrecognizable context.

On multiple mainstream benchmarks, ZoomClick enables state-of-the-art (SOTA) models to achieve substantial breakthroughs, and even allows smaller models to surpass their previously larger counterparts.

\textit{Insight 2: Zoom is far from saturated, but current benchmarks are unable to expose its essential shortcomings.} We present \textbf{GUIZoom-Bench}, which collects evaluation data based on two principles--the zoom depth UI-Venus-72B first hit the target and whether a  prediction maintains after subsequent zooming. We  samples into five types: easy-normal, hard-normal, easy-mislead, hard-mislead and hard-est. Our results further reveals the unresolved headroom of zoom in complex layouts, fine-grained elements, and resolution-mismatch scenarios, offering interpretable standards to design more robust and generalizable zoom methods.

Overall, our contributions can be summarized as follows:
\begin{itemize}
    \item A training-free GUI grounding method that integrates four important capabilities of zoom (\ie, pre-zoom, depth, shrink size, and minimal crop size) within a three-stage pipeline.
    \item A benchmark for evaluating the zoom capabilities of different GUI grounding models, delivering a more specific and interpretable quantitative basis for future zoom improvements.
    \item Experiments demonstrate that ZoomClick enables existing models to achieve state-of-the-art performance in GUI grounding tasks, with smaller models attaning results comparable to those of larger models, \eg, UI-Venus-72B attains a 73.1\% success rate on ScreenSpot-Pro and UI-Venus-7B with ZoomClick outperforms orignal UI-Venus 72B by 2.2\%.
\end{itemize}

\section{Related Works}

\begin{table*}[h]
\centering
\vspace{-5mm}
\scriptsize
\setlength{\tabcolsep}{3pt}
\renewcommand{\arraystretch}{1.05}
\resizebox{\linewidth}{!}{
\begin{tabular}{p{2.9cm}|c|c|c|c|c|c}
\toprule
\textbf{Method} &
\textbf{Training-Free} &
\textbf{Pre-Zoom }{(Sec.\ref{term:prezoom})} &
\textbf{Multi-Step Zoom} (Sec. \ref{term:multi-step-zoom}) &
\textbf{Plain Input} &
\textbf{Plain Output} &
\textbf{Adaptive Context Retention} {(Sec. \ref{term:min_crop_size})} \\
\midrule

GUI-SPOTLIGHT\cite{lei2025textscguispotlightadaptiveiterativefocus} & \xmark & \xmark  & \cmark & \xmark & \xmark & \xmark \\[2pt]

GUI-Cursor\cite{zhao2025learningguigroundingspatial}  & \xmark & \xmark &  \xmark & \xmark & \xmark & \xmark \\[2pt]

R-VLM & \cmark* & \xmark &  \cmark & \cmark & \xmark & \xmark\\[2pt] 

ReGUIDE\cite{lee2025reguidedataefficientgui}& \xmark & \xmark* &  \xmark & \cmark & \cmark & \xmark\\[2pt] 

DiMo-GUI\cite{wu2025dimoguiadvancingtesttimescaling} & \cmark & \xmark &  \cmark & \cmark & \cmark & \xmark \\[2pt]

GMS\cite{li2025generalistscannermeetsspecialist}
& \cmark & \xmark &  \cmark & \xmark & \xmark & \xmark \\[2pt]

RegionFocus\cite{luo2025visualtesttimescalinggui}  & \cmark & \xmark & \cmark & \xmark & \xmark & \xmark \\[2pt]

\rowcolor{gray!10}
\textbf{Ours (ZoomClick)} & \cmark & \cmark & \cmark & \cmark & \cmark & \cmark \\

\bottomrule
\end{tabular}
}
\caption{\textbf{Feature comparison of recent GUI grounding works.} 
\textit{Plain Input} denotes that no additional visual context beyond the raw screenshot is provided; 
\textit{Plain Output} denotes that no VLM-as-judge or external reranking module is used. 
* in the second column means that R-VLM can be deployed without training at inference time.  
* in the third column means that ReGUIDE take global-local consistency into consideration, but different from prezoom. }
\label{tab:gui_feature_comparison}
\end{table*}

\subsection{GUI Agents}
GUI agent autonomously selects actions to accomplish user-specified tasks \cite{chen2024cohtmlagent}\cite{ding2023mind2web}\cite{yin2024browseruse}\cite{zhou2024webarena} by perceiving the interactive interface. Early approaches rely on structured text extracted from GUI environments, such as HTML \cite{ding2023mind2web}\cite{lu2024omniparserpurevisionbased}\cite{zhou2024webarena} or accessibility trees, and utilize large language models (LLMs) to process them\cite{liu2023visualinstructiontuning}. To better understand visual rendering and layout variability, recent approaches have shifted toward MLLMs\cite{guo2025seed15vltechnicalreport} to directly operate on screenshots. Specifically, modular approaches adopts vision–language models (VLMs) as high-level semantic planners\cite{luo2025visualtesttimescalinggui}\cite{wu2025dimoguiadvancingtesttimescaling}\cite{yang2025gta1guitesttimescaling, liu2023llavapluslearningusetools}, complemented by specialized grounding models. In contrast, end-to-end approaches jointly train planning and localization capabilities within a unified framework. Post-training methods are further introduced to enhance reasoning and adaptability, \eg reinforcement learning (RL) \cite{tang2025guig2gaussianrewardmodeling}\cite{zhou2025guig1understandingr1zeroliketraining}.

\subsection{GUI Grounding}
GUI grounding maps natural language instructions to on-screen visual elements\cite{wu2025guiactorcoordinatefreevisualgrounding}, serving as a prerequisite for the planning and action of GUI agents. Existing approaches typically formulate the task as coordinate prediction and finetune VLMs \cite{gu2025uivenustechnicalreportbuilding}\cite{lee2025reguidedataefficientgui}\cite{qin2025uitarspioneeringautomatedgui}\cite{park2025rvlmregionawarevisionlanguage} or general grounding models to leverage their perceptual priors. This paradigm has led to the development of specialized GUI grounding models\cite{zhou2025guiaimaaligningintrinsicmultimodal}, large-scale screenshot–text–position paired datasets\cite{cheng2024seeclickharnessingguigrounding}\cite{li2025screenspotproguigroundingprofessional}\cite{xie2024osworldbenchmarkingmultimodalagents} and corresponding collection/synthesis pipelines, as well as standardized evaluation benchmarks. Notably, zoom has demonstrated strong effectiveness in many visual grounding tasks, and recent works have attempted to introduce this mechanism into above models. R-VLM\cite{park2025rvlmregionawarevisionlanguage} leverages a two-step coarse-to-fine zoom in, followed by an IoU-aware weighted loss to ensure high IoU grounding. DiMo-GUI\cite{wu2025dimoguiadvancingtesttimescaling} decouples layouts into texts and icons, maintaining a search path for each modality, and select between the two candidates based on model judgement. ReGUIDE \cite{lee2025reguidedataefficientgui} adopts a statistics-driven strategy, using Kernel Density Estimation to determine the optimal crop region for robust majority voting at test time. Other methods such as GUI-Cursor\cite{zhao2025learningguigroundingspatial} and RegionFocus\cite{luo2025visualtesttimescalinggui} maintain a single search path with history actions, and relies on the model to determine the next action. These methods leverage zoom either as an auxiliary tool for training or as a simple rule-based heuristic at inference time. As a result, its full potential remains underdeveloped, and a notable performance gap persists between these practices and what zoom can really achieve. Our work close this gap through a comprehensive analysis of zoom behaviors, revealing the principles required to fully unlock its capability.

\section{Zoom in, Click out : A Practical Paradigm for Proper Zoom-in Grounding}
We introduce \method, a training-free test-time search strategy. 
\method uses a fixed shrink ratio together with a minimum crop size and simple boundary handling, which preserves
context and stabilizes narrowing without per-model parameter adjustment. 
This design works consistently across different grounding models and leads to more reliable multi-step localization.

\subsection{Problem Setup}
Given a high-resolution GUI image $I\in\mathbb{R}^{H\times W\times 3}$ and a language query $q$,
a grounding model $\mathcal{G}$ outputs a normalized point $\hat{p}=(\hat{x},\hat{y})\in[0,1]^2$ with respect
to its input view. We maintain a viewport $V=(v_x^1,v_y^1,v_x^2,v_y^2)\in[0,1]^4$ that specifies a region
of $I$ in normalized coordinates. The predicted point in the original image coordinates is
\begin{equation}
p_r=\big(v_x^1+(v_x^2{-}v_x^1)\hat{x},\;\; v_y^1+(v_y^2{-}v_y^1)\hat{y}\big),
\end{equation}
and the corresponding pixel location is
\begin{equation}
p_{\text{px}}=\big(\mathrm{round}(W\cdot p_r^{x}),\; \mathrm{round}(H\cdot p_r^{y})\big).
\end{equation}
We start with the full-image viewport $V_0=(0,0,1,1)$ and iteratively update the viewport during zooming;
all updates operate on the same mapping above.

\subsection{ZoomClick}
\subsubsection{When to Zoom in}
According to our studies in Fig \ref{fig:depth_curves}, great first-step accuracy leads to massive performance boost in latter steps. Therefore, we introduce \textbf{Pre-Zoom}\label{term:prezoom}, a single-shot agreement test between the full image and 
$K$ non-overlapping patches (we use $K{=}4$ for a $2{\times}2$ grid) in the first iteration, as an effective way to \textit{ensure a reliable start}. Let
\[
p^{\text{dir}}=\mathcal{G}(I,q), \qquad
p^{(k)}=\mathcal{G}(I^{(k)},q),
\]
denote the predictions on the full image and the $k$-th patch, respectively, each mapped back to the
original image coordinates.
We compute pixel-space distances
\[
d_k=\|p^{\text{dir}} - p^{(k)}\|_2, \qquad
k^\star = \arg\min_k d_k,
\]
and use a threshold $\tau$ (in pixels) to select the initial point:
\[
p^{(1)}=
\begin{cases}
p^{(k^\star)}, & d_{k^\star}<\tau,\\[3pt]
p^{\text{dir}}, & \text{otherwise}.
\end{cases}
\]
\noindent{Intuitively, if any tile closely agrees with the global prediction, it provides a \emph{cleaner local context} 
for initiating zoom; otherwise, we retain the global view to avoid premature narrowing.}

\begin{figure*}
    \centering
    \includegraphics[width=\linewidth]{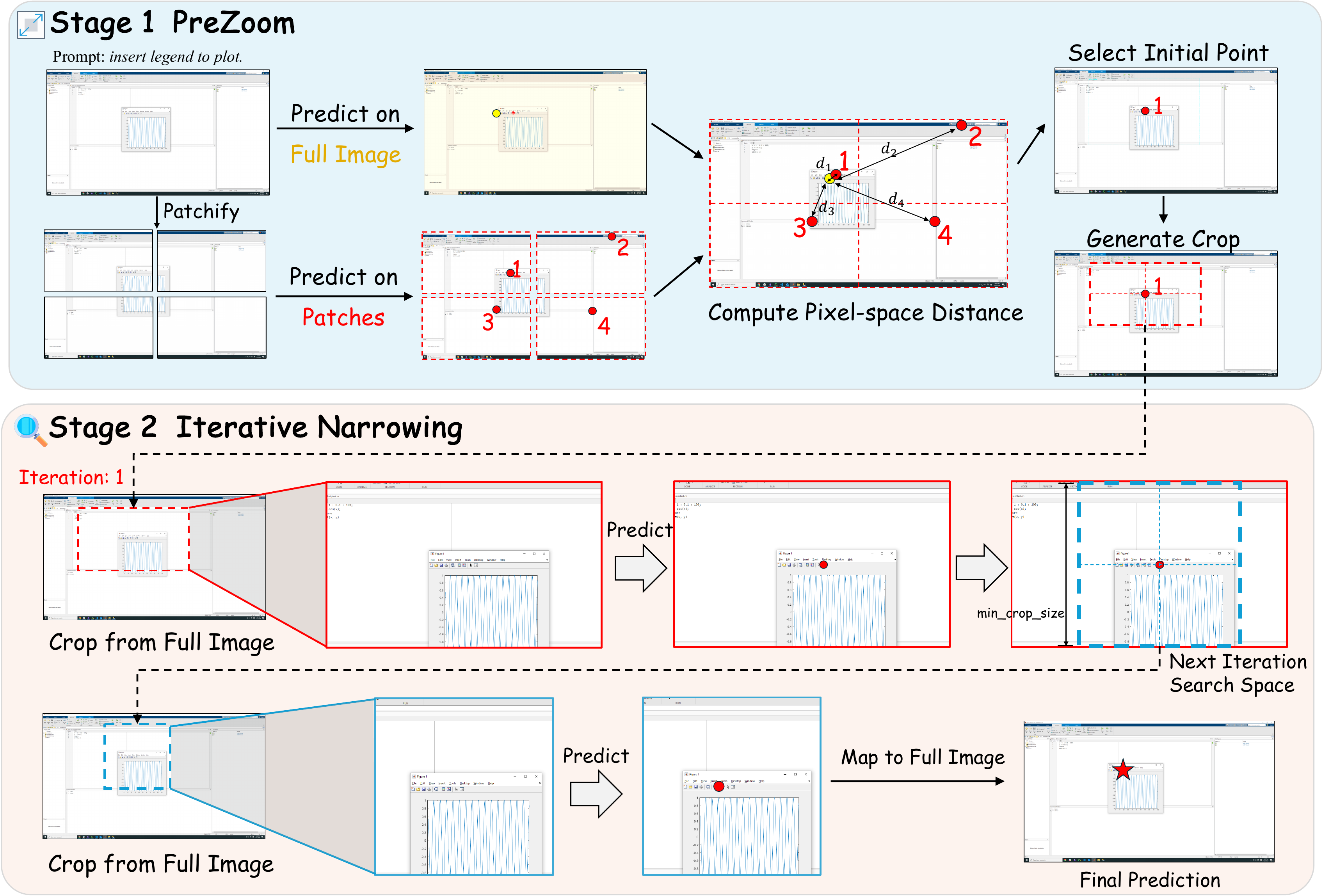}
    \caption{\textbf{Model Framework of ZoomClick.}  \text{Min\_crop\_size} \ref{term:min_crop_size} represents the lower bound of viewport during Iterative Narrowing.}
    \label{fig:placeholder}
\end{figure*}

\subsubsection{How to Zoom in}
\textbf{Multi-step Narrow-in}\label{term:multi-step-zoom} Instead of a one-shot crop, \method{} progressively navigates and refines the viewport around each prediction. This gradual shrinking enables the model to iteratively correct localization errors and converge to the precise target.  

At each iteration $t$, \method updates a localized view by shrinking the current viewport around the predicted point.
Given viewport size $(W_t, H_t)$ and prediction $p^{(t)}$, we compute a new crop size:
\[
(\tilde{W},\tilde{H}) = \big(\max(\lfloor \rho W_t \rfloor, m),\ \max(\lfloor \rho H_t \rfloor, m)\big),
\]
where $\rho\in(0,1)$ is a fixed shrink ratio and $m$ is a minimum crop size that preserves essential context.


\subsubsection{When to Click out}
\textbf{Minimum Crop Size}\label{term:min_crop_size}
The minimum crop size $m$ acts as a lower bound that prevents the view from collapsing into an overly tight and noisy region.
Formally, the effective zoom level at iteration $t$ satisfies:
\[
\min(W_t, H_t) \ge m,
\]
\noindent
ensuring that a non-trivial spatial neighborhood around $p^{(t)}$ is always preserved. This \emph{context floor} keeps the view within the model’s training resolution regime and retains semantic cues (e.g., surrounding text, icon grouping, layout)
needed for reliable grounding.

\noindent
\textbf{Viewport Boundary Handling}
If a crop exceeds the image boundary, we apply:
\texttt{shift} (move window inside),
\texttt{clip} (trim overflow),
or \texttt{shrink} (reduce window while keeping $p^{(t)}$ centered).
We use \texttt{shift} by default; \texttt{clip} is useful when the prediction is already accurate and further motion is unnecessary.

\noindent
\textbf{Termination} We stop after $T$ iterations or once the crop reaches the minimum size. The final click is the last mapped prediction
$p_r^{(T)}$, converted to pixel coordinates $p_{\text{px}}$.

\section{GUIZoom-Bench: Benchmarking the Behavioral Patterns of Zoom-in Strategies}


\begin{figure*}[t]
\vspace{-5mm}
    \centering
    \includegraphics[width=\linewidth]{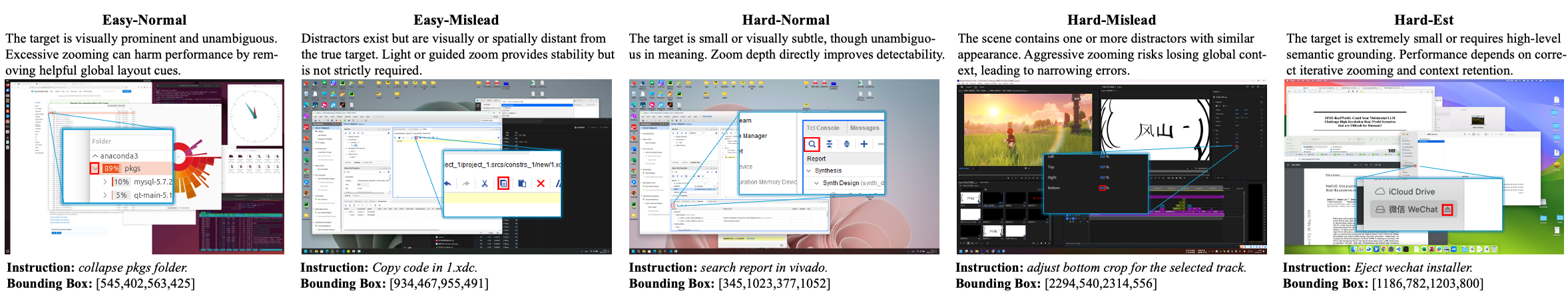}
    \caption{\textbf{Data examples of each category in GUIZoom-Bench.}}
    \label{fig:placeholder}
\end{figure*}

The potential of zoom is far from saturated, yet existing GUI grounding benchmarks fail to reveal its essential shortcomings or the model-specific limitations that emerge under different zoom conditions. Current evaluations do not clarify when zoom improves localization, when it induces harmful context loss or distraction, nor why certain models collapse under deeper zoom levels—leaving little guidance for advancing zoom-aware methods. To address this gap, we introduce \textbf{GUIZoom-Bench}, a benchmark designed to systematically dissect zoom behavior, exposing both the failure modes of zoom itself and the adaptability boundaries of different model families. These insights not only pinpoint the true sources of performance gains but also provide concrete directions for developing more effective and robust zoom-aware grounding models.

\begin{figure}
    \centering
    \includegraphics[width=\linewidth]{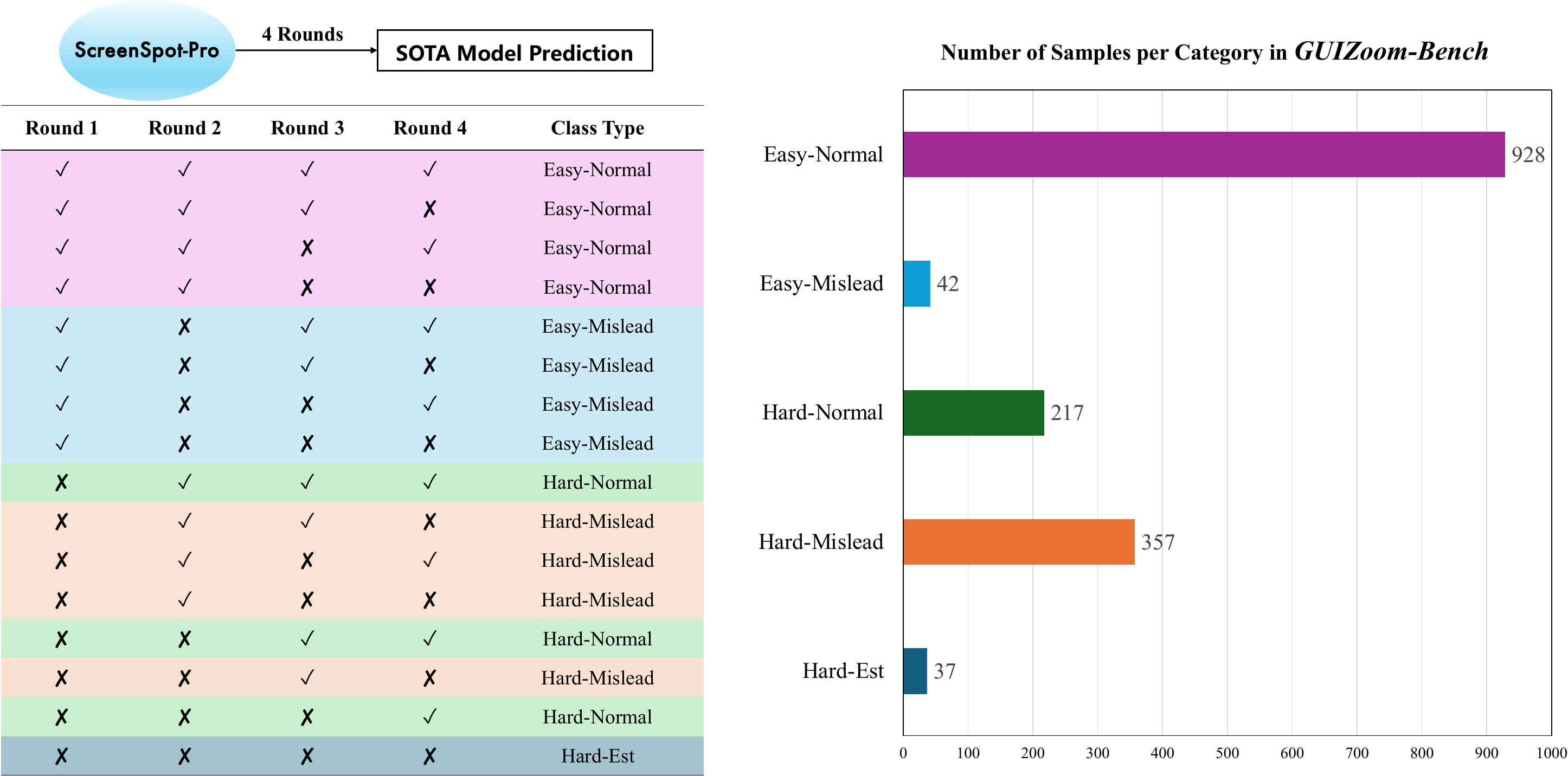}
    \caption{\textbf{Data organization in our proposed GUIZoom-Bench.}}
    \label{fig:difficulty_reliability}
\end{figure}

\subsection{Five Zoom Behaviors: Difficulty × Reliability}
Zoom does not benefit all samples in the same way. Some targets are easily recognized without zooming, others gradually emerge through progressive narrowing, and some are even misled by lost context or visual distractors. \textbf{GUIZoom-Bench} captures these behavioral differences by partitioning samples into five subsets that reveal \textit{how} zoom shapes model perception—highlighting when it aids, when it distracts, and where it reaches its limits.


\begin{itemize}[leftmargin=1.6em]
\item \textit{\textbf{easy\_normal.}} The target is visually salient and contextually clear; In this case, \emph{excessive zooming can hurt} by unnecessarily discarding useful context.

\item \textbf{\textit{easy\_mislead.}} The model starts correct but becomes wrong after zooming, as distractors grow more salient once global cues vanish. This reflects a \emph{loss of stability} caused by over-focusing.

\item \textbf{\textit{hard\_normal.}} The target is small or visually subtle, initially hard to locate but gradually revealed through zooming. This showcases the \emph{productive role} of zoom in exposing fine-grained details.

\item \textbf{\textit{hard\_mislead.}} The scene contains distractors visually similar to the target, making zooming prone to shift attention away. These cases capture \emph{context-induced confusion} under narrowing.

\item \textbf{\textit{hard\_est.}} Even repeated zooming fails to clarify the target due to extreme subtlety or vague instruction. This subset defines the \emph{upper limit} of zoom-enabled grounding.
\end{itemize}

\noindent
Together, these five categories delineate a two-dimensional space of \textbf{difficulty × reliability}, where \textit{difficulty} captures the inherent visual and contextual complexity of a target, and \textit{reliability} reflects the model’s behavioral stability under zooming.


\subsection{Collection Method and Criteria}



We construct \textbf{GUIZoom-Bench} by reorganizing samples from \textit{ScreenSpot-Pro}, a large-scale GUI grounding dataset with high-resolution interfaces and dense, fine-grained targets—making it a \textit{natural testbed} for analyzing zoom behavior.

To capture how zoom affects grounding dynamics, we run our state-of-the-art \textit{\method{}} model (UI-Venus-72B) under a fixed four-step zoom-in procedure and record correctness at each iteration. Each ScreenSpot-Pro sample is evaluated under four zoom-in rounds using the SOTA grounding model, producing a correctness sequence $\{s_1,s_2,s_3,s_4\}$, where $s_t{=}1$ indicates a correct click at iteration $t$.

From these correctness sequences, we extract two behavioral factors that define the taxonomy shown in Fig. \ref{fig:difficulty_reliability} (Left). The \textit{iteration of first correctness} quantifies the sample’s \textbf{difficulty}—how many zoom steps are needed before the model finds the target—while the \textit{stability of correctness} reflects \textbf{reliability}—whether the model correctness remains once zoomed in.

By crossing these two dimensions, we categorize every sample into one of five behavioral types (\textit{easy\_normal, easy\_mislead, hard\_normal, hard\_mislead, hard\_est}). The number of samples per category is shown in Fig. \ref{fig:difficulty_reliability} (Right).

\begin{figure*}
    \centering
    \includegraphics[width=\linewidth]{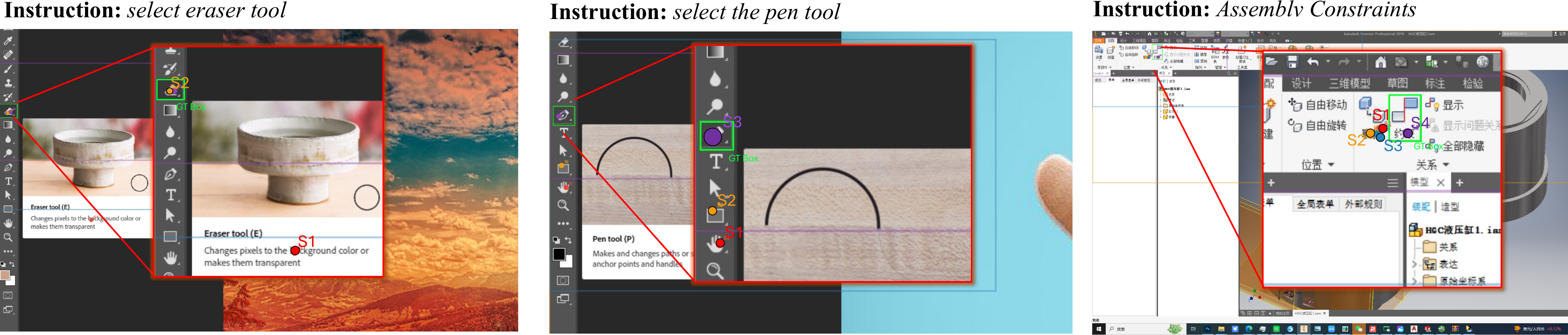}
\caption{\textbf{Qualitative results of \method{} on ScreenSpot-Pro.} Left: depth-2 zoom resolves the error from depth 1. Center: depth-3 zoom resolves the errors from depths 1–2. Right: depth-4 zoom resolves the errors from depths 1–3.}
    \label{fig:qualitative_sspro}
\end{figure*}

\section{Experiments}
\subsection{Experimental Setup}

\noindent
\textbf{Model Selection} We evaluate \method{} on two strong GUI grounding models with automatic localization:
\textit{Qwen3-VL}\cite{yang2025qwen3technicalreport} (general-purpose VLM) and \textit{UI-Venus} (GUI-specific multimodal agent). Together, they span open-domain reasoning and specialized UI interaction, allowing us to test \method{}’s generalization across paradigms under a unified zoom inference schedule.


\vspace{0.5em}
\noindent
\textbf{Datasets and Benchmarks}
We evaluate on two complementary benchmarks for realistic desktop-scale GUIs. \textit{ScreenSpot-Pro} provides high-resolution professional interfaces with densely packed fine-grained elements, ideal for testing visual precision under zoom. \textit{UI-Vision}\cite{nayak2025uivisiondesktopcentricguibenchmark} includes diverse real-world software annotated with elements and interaction traces, emphasizing functional understanding and realistic interaction contexts. Together, they combine visual complexity and interaction diversity, forming a natural testbed for assessing the robustness and effectiveness of zoom-based grounding.

\vspace{0.5em}
\noindent
\textbf{Metric}
We report grounding accuracy—the proportion of correctly localized targets—under different zoom behaviors and datasets, following the official evaluation setup of each benchmark to ensure fair comparison. Full implementation details can be found in Appendix A.

\subsection{Main Results}
\subsubsection{ScreenSpot-Pro}
\method{} yields \textit{consistent and significant improvements} of grounding accuracy over base models. As shown in Table \ref{tab:sspro},  \method{} + Qwen3-VL-32B achieves a \textbf{34.4\%} relative improvement over the base model, Qwen3-VL-32B. A similar trend holds for GUI-specific models: \method{} + UI-Venus-7B achieves 30.6\% relative accuracy gain, surpassing even the base UI-Venus-72B model. Moreover, even large and strong backbones benefit from \method{},  with  UI-Venus-72B gaining an 19.1\% accuracy improvement, setting a new record of \textbf{73.1\%} for ScreenSpot-Pro. 

\begin{table*}[t]
\vspace{-5mm}
\centering
\small
\setlength{\tabcolsep}{3pt}
\resizebox{\textwidth}{!}{
\begin{tabular}{l|ccc|ccc|ccc|ccc|ccc|ccc|ccc}
\toprule
\textbf{Methods} &
\multicolumn{3}{c|}{\textbf{Development}} &
\multicolumn{3}{c|}{\textbf{Creative}} &
\multicolumn{3}{c|}{\textbf{CAD}} &
\multicolumn{3}{c|}{\textbf{Scientific}} &
\multicolumn{3}{c|}{\textbf{Office}} &
\multicolumn{3}{c|}{\textbf{OS}} &
\multicolumn{3}{c}{\textbf{Avg}}\\
 & text & icon & avg & text & icon & avg & text & icon & avg & text & icon & avg & text & icon & avg & text & icon & avg & text & icon & avg \\
\midrule
\SectionRow{Proprietary Methods}
GPT-4o\cite{openai2024gpt4ocard} & 1.3 & 0.0 & 0.7 & 1.0 & 0.0 & 0.6 & 2.0 & 0.0 & 1.5 & 2.1 & 0.0 & 1.2 & 1.1 & 0.0 & 0.9 & 0.0 & 0.0 & 0.0 & 1.3 & 0.0 & 0.8 \\[2pt]
Claude-3.7-Sonnet\cite{anthropic2025claude37} & - & - & - & - & - & - & - & - & - & - & - & - & - & - & - & - & - & - & - & - & 27.7 \\[2pt]
Seed-1.5-VL\cite{guo2025seed15vltechnicalreport} & - & - & 53.8 & - & - & 59.2 & - & - & 59.0 & - & - & 61.4 & - & - & 74.8 & - & - & 60.2 & - & - & 60.9 \\[2pt]
UI-TARS-1.5\cite{qin2025uitarspioneeringautomatedgui} & - & - & 63.9 & - & - & 50.4 & - & - & 58.2 & - & - & 69.3 & - & - & 79.6 & - & - & 51.0 & - & - & 61.6 \\[2pt]
\midrule
\SectionRow{Open-Sourced Methods}
OS-Atlas-7B\cite{wu2024osatlasfoundationactionmodel} & 33.1 & 1.4 & 17.7 & 28.8 & 2.8 & 17.9 & 12.2 & 4.7 & 10.3 & 37.5 & 7.3 & 24.4 & 33.9 & 5.7 & 27.4 & 27.1 & 4.5 & 16.8 & 28.1 & 4.0 & 18.9 \\[2pt]
Qwen2.5-VL-3B\cite{bai2025qwen25vltechnicalreport} & 38.3 & 3.4 & 21.4 & 40.9 & 4.9 & 25.8 & 22.3 & 6.3 & 18.4 & 44.4 & 10.0 & 29.5 & 48.0 & 17.0 & 40.9 & 33.6 & 4.5 & 20.4 & 37.8 & 6.6 & 25.9 \\[2pt]
Qwen2.5-VL-7B & 51.9 & 4.8 & 29.1 & 36.9 & 8.4 & 24.9 & 17.8 & 1.6 & 13.8 & 48.6 & 8.2 & 31.1 & 53.7 & 18.9 & 45.7 & 34.6 & 7.9 & 22.4 & 39.9 & 7.6 & 26.8 \\[2pt]
UGround-7B\cite{gou2025navigatingdigitalworldhumans} & - & - & 35.5 & - & - & 27.8 & - & - & 13.5 & - & - & 38.8 & - & - & 48.8 & - & - & 26.1 & - & - & 31.1 \\[2pt]
UGround-72B & - & - & 31.1 & - & - & 35.8 & - & - & 13.8 & - & - & 50.0 & - & - & 51.3 & - & - & 25.5 & - & - & 34.5 \\[2pt]
UI-TARS-7B & 58.4 & 12.4 & 36.1 & 50.0 & 9.1 & 32.8 & 20.8 & 9.4 & 18.0 & 63.9 & 31.8 & 50.0 & 63.3 & 20.8 & 53.5 & 30.8 & 16.9 & 24.5 & 47.8 & 16.2 & 35.7 \\[2pt]
UI-TARS-72B & 63.0 & 17.3 & 40.8 & 57.1 & 15.4 & 39.6 & 18.8 & 12.5 & 17.2 & 64.6 & 20.9 & 45.7 & 63.3 & 26.4 & 54.8 & 42.1 & 15.7 & 30.1 & 50.9 & 17.5 & 38.1 \\[2pt]
Jedi-7B\cite{xie2025scalingcomputerusegroundinguser} & 42.9 & 11.0 & 27.4 & 50.0 & 11.9 & 34.0 & 38.0 & 14.1 & 32.2 & 72.9 & 25.5 & 52.4 & 75.1 & 47.2 & 68.7 & 33.6 & 16.9 & 26.0 & 52.6 & 18.2 & 39.5 \\[2pt]
Qwen2.5-VL-32B & 74.0 & 21.4 & 48.5 & 61.1 & 13.3 & 41.1 & 38.1 & 15.6 & 32.6 & 78.5 & 29.1 & 57.1 & 76.3 & 37.7 & 67.4 & 55.1 & 27.0 & 42.3 & 63.2 & 22.5 & 47.6 \\[2pt]
GTA1-7B\cite{yang2025gta1guitesttimescaling} & 53.3 & 17.2 & 44.5 & 66.9 & 20.7 & 44.0 & 62.6 & 18.2 & 44.4 & 76.4 & 31.8 & 57.1 & 82.5 & 50.9 & 75.2 & 48.6 & 25.9 & 38.3 & 65.5 & 25.2 & 50.1 \\[2pt]
GTA1-32B & 83.1 & 37.9 & 61.2 & 72.2 & 25.9 & 52.8 & 70.1 & 31.3 & 60.5 & 84.7 & 39.1 & 65.0 & 89.3 & 64.2 & 83.5 & 76.6 & 51.7 & 65.3 & 78.9 & 38.9 & 63.6 \\[2pt]
\midrule
\SectionRow{Ours}

\multirow{1}{*}{Qwen3-VL-32B} 
& 74.0 & 13.1 & 44.5 
& 74.2 & 23.1 & 52.8 
& 60.4 & 23.4 & 51.3 
& 78.5 & 30.9 & 57.9 
& 81.9 & 41.5 & 72.6 
& 66.4 & 24.7 & 47.4 
& 72.6 & 24.0 & 54.0 \\

\multirow{1}{*}{\textit{+ ZoomCLick}} 
& \textbf{88.3} & \textbf{55.9} & \textbf{72.6} 
& \textbf{82.8} & \textbf{38.5} & \textbf{64.2} 
& \textbf{80.7} & \textbf{42.2} & \textbf{71.3} 
& \textbf{90.2} & \textbf{40.9} & \textbf{68.9} 
& \textbf{93.8} & \textbf{77.4} & \textbf{90.0} 
& \textbf{84.1} & \textbf{51.7} & \textbf{69.4} 
& \textbf{86.5} & \textbf{48.8} & \textbf{72.1} \\

\multirow{1}{*}{\textit{$\Delta$}} 
& \textcolor{red}{+14.3} & \textcolor{red}{+42.8} & \textcolor{red}{+28.1} 
& \textcolor{red}{+8.6} & \textcolor{red}{+15.4} & \textcolor{red}{+11.4} 
& \textcolor{red}{+20.3} & \textcolor{red}{+18.8} & \textcolor{red}{+20.0} 
& \textcolor{red}{+11.7} & \textcolor{red}{+10.0} & \textcolor{red}{+11.0} 
& \textcolor{red}{+11.9} & \textcolor{red}{+35.9} & \textcolor{red}{+17.4} 
& \textcolor{red}{+17.7} & \textcolor{red}{+27.0} & \textcolor{red}{+22.0} 
& \textcolor{red}{+13.9} & \textcolor{red}{+24.8} & \textcolor{red}{+18.1} \\[4pt]

\multirow{1}{*}{UI-Venus-7B}
& 73.7 & 24.1 & 49.8 
& 61.6 & 14.7 & 41.9 
& 58.9 & 18.8 & 49.0 
& 76.4 & 33.6 & 57.9 
& 75.1 & 43.4 & 67.8 
& 49.5 & 22.5 & 37.2 
& 66.3 & 24.5 & 50.3 \\

\multirow{1}{*}{\textit{+ ZoomCLick}} 
& \textbf{83.1} & \textbf{45.5} & \textbf{64.9}
& \textbf{74.2} & \textbf{30.8} & \textbf{56.0} 
& \textbf{79.2} & \textbf{32.8} & \textbf{67.8} 
& \textbf{86.8} & \textbf{42.7} & \textbf{67.7} 
& \textbf{88.7} & \textbf{52.8} & \textbf{80.4} 
& \textbf{78.5} & \textbf{39.3} & \textbf{60.7} 
& \textbf{81.6} & \textbf{39.9} & \textbf{65.7} \\

\multirow{1}{*}{\textit{$\Delta$}} 
& \textcolor{red}{+9.1} & \textcolor{red}{+21.4} & \textcolor{red}{+15.1} 
& \textcolor{red}{+12.6} & \textcolor{red}{+16.1} & \textcolor{red}{+14.1} 
& \textcolor{red}{+20.3} & \textcolor{red}{+14.0} & \textcolor{red}{+18.8} 
& \textcolor{red}{+10.4} & \textcolor{red}{+9.1} & \textcolor{red}{+9.8} 
& \textcolor{red}{+13.6} & \textcolor{red}{+9.4} & \textcolor{red}{+12.6} 
& \textcolor{red}{+29.0} & \textcolor{red}{+16.8} & \textcolor{red}{+23.5} 
& \textcolor{red}{+15.3} & \textcolor{red}{+15.4} & \textcolor{red}{+15.4} \\[4pt]

\multirow{1}{*}{UI-Venus-72B}
& 82.5 & 33.8 & 58.9 
& 73.7 & 30.8 & 55.7 
& 65.5 & 29.7 & 56.7 
& 84.0 & 42.7 & 66.1 
& 82.5 & 58.5 & 77.0 
& 73.8 & 36.0 & 56.6 
& 76.6 & 36.8 & 61.4 \\

\multirow{1}{*}{\textit{+ ZoomCLick}} 
& \textbf{88.3} & \textbf{51.0} & \textbf{70.2} 
& \textbf{81.8} & \textbf{49.7} & \textbf{68.3} 
& \textbf{85.3} & \textbf{48.4} & \textbf{76.2} 
& \textbf{91.0} & \textbf{48.1} & \textbf{72.4} 
& \textbf{91.5} & \textbf{66.0} & \textbf{85.7} 
& \textbf{79.4} & \textbf{52.8} & \textbf{67.3} 
& \textbf{86.4} & \textbf{51.5} & \textbf{73.1} \\

\multirow{1}{*}{\textit{$\Delta$}} 
& \textcolor{red}{+5.8} & \textcolor{red}{+17.2} & \textcolor{red}{+11.3} 
& \textcolor{red}{+8.1} & \textcolor{red}{+18.9} & \textcolor{red}{+12.6} 
& \textcolor{red}{+19.8} & \textcolor{red}{+18.7} & \textcolor{red}{+19.5} 
& \textcolor{red}{+7.0} & \textcolor{red}{+5.4} & \textcolor{red}{+6.3} 
& \textcolor{red}{+9.0} & \textcolor{red}{+7.5} & \textcolor{red}{+8.7} 
& \textcolor{red}{+5.6} & \textcolor{red}{+16.8} & \textcolor{red}{+10.7} 
& \textcolor{red}{+9.8} & \textcolor{red}{+14.7} & \textcolor{red}{+11.7} \\

\bottomrule
\end{tabular}
}
\caption{\textbf{Comparison of Methods on ScreenSpot-Pro.} }
\label{tab:sspro}
\end{table*}

\subsubsection{UI-Vision}
As shown in Table \ref{tab:uivision} , \method{} delivers a substantial improvement over prior work on UI-Vision. When applied to UI-Venus-72B, it raises click accuracy from \textbf{25.5\%} to \textbf{42.5\%} (+66.7\% relative), establishing a new state of the art. These gains persist across both \textit{model scales }(7B–72B) and \textit{model types} (UI-specific vs. general-purpose VLMs), demonstrating that our training-free refinement reliably enhances grounding across architectures rather than relying on model or dataset specialization.

\noindent
\textbf{Analysis}
The effectiveness of \method{} stems from its simplicity: a training-free procedure that \emph{leverages the strong priors already embedded in modern VLMs and grounding-specific models}. These models are generally capable of identifying semantically plausible regions, but often lack the mechanisms to translate coarse alignment into precise localization. \method{} exposes and amplifies this through a minimal multi-step zooming routine, without introducing additional learning or architectural changes. Despite its simplicity, this lightweight refinement reliably converts existing semantic priors into state-of-the-art grounding performance on high-resolution desktop interfaces.


\begin{table}[t] 
\centering
\scriptsize
\setlength{\tabcolsep}{2pt}
\renewcommand{\arraystretch}{0.9}

\resizebox{0.9\columnwidth}{!}{%
\begin{tabular}{l|c|c|c|c}
\toprule
\multirow{2}{*}{\textbf{Method}} &
\multicolumn{1}{c|}{\textbf{Basic}} &
\multicolumn{1}{c|}{\textbf{Func}} &
\multicolumn{1}{c|}{\textbf{Spatial}} &
\multirow{2}{*}{\textbf{Final Avg}}\\
& Overall & Overall & Overall & \\
\midrule
\rowcolor{gray!12}\multicolumn{5}{l}{\textit{Closed-Source VLMs}}\\
GPT\mbox{-}4o & 1.6 & 1.5 & 1.0 & 1.4\\
Gemini\mbox{-}1.5\mbox{-}pro\cite{geminiteam2024gemini15unlockingmultimodal} & 0.8 & 0.3 & 0.6 & 0.6\\
Claude\mbox{-}3.7\mbox{-}Sonnet & 9.5 & 7.7 & 7.6 & 8.3\\
\midrule
\rowcolor{gray!12}\multicolumn{5}{l}{\textit{Open-Source VLMs}}\\
Qwen\mbox{-}2.5VL\mbox{-}7B & 1.2 & 0.8 & 0.5 & 0.9\\
InternVL2.5\mbox{-}8B\cite{chen2025expandingperformanceboundariesopensource} & 2.5 & 2.8 & 1.0 & 2.1\\
Qwen\mbox{-}2VL\mbox{-}7B\cite{wang2024qwen2vlenhancingvisionlanguagemodels} & 3.4 & 3.2 & 1.5 & 2.7\\
MiniCPM\mbox{-}V\mbox{-}8B\cite{yao2024minicpmvgpt4vlevelmllm} & 7.1 & 5.3 & 1.5 & 4.3\\
\midrule
\rowcolor{gray!12}\multicolumn{5}{l}{\textit{Open-Source GUI Agents}}\\
ShowUI\mbox{-}2B\cite{lin2024showuivisionlanguageactionmodelgui} & 8.1 & 7.7 & 2.1 & 5.9\\
AriaUI\mbox{-}25.3B\cite{yang2025ariauivisualgroundinggui} & 12.2 & 14.0 & 4.0 & 10.1\\
UGround\mbox{-}v1\mbox{-}7B & 15.4 & 17.1 & 6.3 & 12.9\\
OSAtlas\mbox{-}7B & 12.2 & 11.2 & 3.7 & 9.0\\
UGround\mbox{-}7B & 11.5 & 12.2 & 2.8 & 8.8\\
Aguvis\mbox{-}7B\cite{xu2025aguvisunifiedpurevision} & 17.8 & 18.3 & 5.1 & 13.7\\
UI\mbox{-}TARS\mbox{-}7B & 20.1 & 24.3 & 8.4 & 17.6\\
CogAgent\mbox{-}9B\cite{hong2024cogagentvisuallanguagemodel} & 12.0 & 12.2 & 2.6 & 8.9\\
SeeClick\mbox{-}9.6B\cite{cheng2024seeclickharnessingguigrounding} & 9.4 & 4.7 & 2.1 & 5.4\\
UGround\mbox{-}v1\mbox{-}72B & 27.9 & 26.7 & 14.9 & 23.2\\
UI\mbox{-}TARS\mbox{-}72B & 31.4 & 30.5 & 14.7 & 25.5\\
\midrule
\rowcolor{gray!12}\multicolumn{5}{l}{\textit{Ours}}\\
UI\mbox{-}Venus\mbox{-}7B & 34.0 & 30.7 & 10.8 & 25.1\\
\textit{+ZoomClick} & \textbf{43.1} & \textbf{39.9} & \textbf{19.1} & \textbf{34.0}\\
\textit{$\Delta$} & \textcolor{red}{+9.1} & \textcolor{red}{+9.2} & \textcolor{red}{+8.3} & \textcolor{red}{+8.9}\\[2pt]
Qwen3\mbox{-}VL\mbox{-}32B & 44.2 & 42.2 & 24.4 & 36.6\\
\textit{+ZoomClick} & \textbf{45.1} & \textbf{45.1} & \textbf{28.3} & \textbf{39.2}\\
\textit{$\Delta$} & \textcolor{red}{+0.9} & \textcolor{red}{+2.9} & \textcolor{red}{+3.9} & \textcolor{red}{+2.6}\\[2pt]
UI\mbox{-}Venus\mbox{-}72B & 45.6 & 42.3 & 23.7 & 36.8\\
\textit{+ZoomClick} & \textbf{50.2} & \textbf{46.6} & \textbf{31.7} & \textbf{42.5}\\
\textit{$\Delta$} & \textcolor{red}{+4.6} & \textcolor{red}{+4.3} & \textcolor{red}{+8.0} & \textcolor{red}{+5.7}\\
\bottomrule
\end{tabular}
}
\caption{\textbf{Comparison of Methods on UI-Vision.}}
\label{tab:uivision}
\vspace{-2mm}
\end{table}

\vspace{-1mm}
\subsection{Ablations}
Our ablation results reveal a clear and surprisingly consistent pattern: \textit{implementing \method{} with the simplest variant of each component achieves best performance.} These findings underscore a central insight of this work—successful zoom-based grounding comes from simply allowing the model’s inherent localization priors to operate without disruption.

\vspace{-1mm}
\subsubsection{When to Zoom in}\label{subsubsec:prezoom}
\begin{figure}[t]
\centering
\resizebox{0.8\linewidth}{!}{
\begin{tikzpicture}
\begin{axis}[
    xmin=1, xmax=4, ymin=20, ymax=75,
    xtick={1,2,3,4}, xlabel={Depth}, ylabel={Accuracy (\%)},
    grid=both, thick, mark size=2.6pt,
    legend style={font=\small, at={(0.5,1.03)}, anchor=south, legend columns=2,
                  /tikz/every even column/.append style={column sep=6pt}},
]
\addplot+ coordinates {(1,57.4) (2,65.7) (3,67.7) (4,66.6)}; \addlegendentry{Qwen3-VL Instruct (235B-A22B)}
\addplot+ coordinates {(1,26.8) (2,41.2) (3,43.7) (4,43.8)}; \addlegendentry{Qwen-2.5-VL Instruct 7B}
\addplot+ coordinates {(1,52.2) (2,62.4) (3,61.4) (4,61.6)}; \addlegendentry{Qwen-2.5-VL Instruct 72B}
\addplot+ coordinates {(1,32.0) (2,45.6) (3,46.7) (4,46.3)}; \addlegendentry{UI-TARS 7B}
\addplot+ coordinates {(1,42.3) (2,52.2) (3,54.1) (4,54.3)}; \addlegendentry{UI-TARS-1.5 7B}
\addplot+ coordinates {(1,50.3) (2,60.9) (3,63.9) (4,64.1)}; \addlegendentry{UI-Venus 7B}
\addplot+ coordinates {(1,61.4) (2,70.5) (3,72.1) (4,72.7)}; \addlegendentry{UI-Venus 72B}
\end{axis}
\end{tikzpicture}
}
\caption{\textbf{Accuracy vs. zoom depth (1–4) across models.}}
\label{fig:depth_curves}
\end{figure}
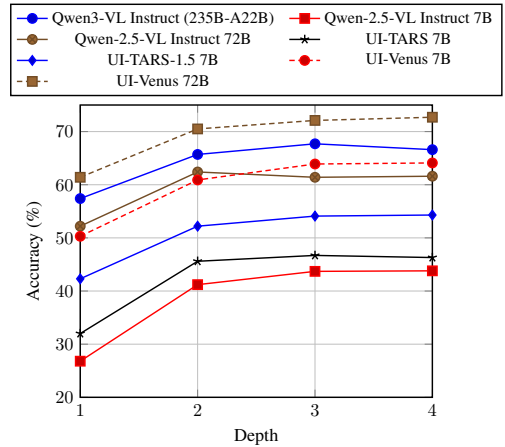

\textit{The importance of Pre-zoom.} As shown in Fig \ref{fig:depth_curves}, it's progressively harder for the model to recover from earlier localization errors, demonstrating the importance of a reliable start. The results from Table \ref{tab:prezoom_compare} proves the effectiveness of Pre-zoom: across all models, Pre-zoom improves grounding accuracy at all depths—for instance, Qwen3-VL-32B rose from \textbf{67.2\%→71.3\%}(\textbf{+4.1\%}) at Depth 2—indicating that early coarse localization provides a stable initialization for subsequent zooming.


\begin{table}[h]
\vspace{0.3em}
\centering
\resizebox{0.9\linewidth}{!}{
\begin{tabular}{l l c c c}
\toprule
\textbf{Model} & \textbf{Prezoom Type} & \textbf{Depth 2} & \textbf{Depth 3} & \textbf{Depth 4} \\
\midrule

\multirow{2}{*}{Qwen3-VL-32B}
& No prezoom              
    & 67.2 & 68.9 & 69.1 \\
& \textbf{Distance-based} 
    & \textbf{71.3} \textcolor{red}{\scriptsize (+4.1)}
    & \textbf{72.1} \textcolor{red}{\scriptsize (+3.2)}
    & \textbf{71.8} \textcolor{red}{\scriptsize (+2.7)} \\
\midrule

\multirow{2}{*}{UI-Venus-7B}
& No prezoom              
    & 60.9 & 63.9 & 64.1 \\
& \textbf{Distance-based} 
    & \textbf{62.4} \textcolor{red}{\scriptsize (+1.5)}
    & \textbf{64.5} \textcolor{red}{\scriptsize (+0.6)}
    & \textbf{64.8} \textcolor{red}{\scriptsize (+0.7)} \\
\midrule

\multirow{2}{*}{UI-Venus-72B}
& No prezoom              
    & 70.5 & 72.1 & 72.7 \\
& \textbf{Distance-based} 
    & \textbf{71.9} \textcolor{red}{\scriptsize (+1.4)}
    & \textbf{73.1} \textcolor{red}{\scriptsize (+1.0)}
    & \textbf{72.9} \textcolor{red}{\scriptsize (+0.2)} \\
\bottomrule
\end{tabular}
}
\vspace{-1mm}
\caption{\textbf{Effect of Pre-zoom across Zoom Depths.}
Accuracy (\%) with/without pre-zoom; red numbers show the absolute gain.}
\label{tab:prezoom_compare}
\vspace{-2mm}
\end{table}

\noindent
\textit{Distance-based Pre-zoom yields better performance.} 
We evaluate two basic approach for Pre-zoom: VLM-based and Distance-based. As shown in Table~\ref{tab:prezoom_vlm_vs_dist}, the \textit{Distance-based} pre-zoom yields consistently higher accuracy—surpassing VLM-based approach on Qwen3-VL-32B by \textbf{9.4\%} and on UI-Venus-72B by 5.1\%.  

\begin{table}[t]
\centering
\vspace{0.3em}
\resizebox{0.9\linewidth}{!}{
\begin{tabular}{l l c c c}
\toprule
\textbf{Model} & \textbf{Pre-zoom Type} & \textbf{Depth 2} & \textbf{Depth 3} & \textbf{Depth 4} \\
\midrule
\multirow{2}{*}{Qwen3-VL-32B}& VLM-based              & 61.9 & 65.5 & 63.4 \\
& \textbf{Distance-based} & \textbf{71.3} & \textbf{72.1} & \textbf{71.8} \\
\midrule
\multirow{2}{*}{UI-Venus-7B}& VLM-based              & 57.7 & 60.9 & 60.9 \\
& \textbf{Distance-based} & \textbf{62.4} & \textbf{64.5} & \textbf{64.8} \\
\midrule
\multirow{2}{*}{UI-Venus-72B}& VLM-based              & 66.8 & 69.0 & 68.6 \\
& \textbf{Distance-based} & \textbf{71.9} & \textbf{73.1} & \textbf{72.9} \\
\bottomrule
\end{tabular}}
\vspace{-1mm}
\caption{\textbf{Pre-zoom Strategy Comparison (VLM-based vs \\Distance-based).} Accuracy (\%) at different zoom depths for each model.}
\label{tab:prezoom_vlm_vs_dist}
\end{table}

\vspace{0.5em}
\noindent
\textbf{Insights}
Findings above suggest a simple but actionable principle for future zoom-based grounding: \textit{intervene only where necessary, and intervene minimally}. A small, well-placed adjustment to the initial region already produces the most meaningful gains, while extra semantic or heuristic reasoning adds little value and can introduce instability. This highlights a clearer direction for future work—before adding external complexity to the method, make full use of the intrinsic potential of zoom itself.

\subsubsection{How to Zoom in}
\textit{The effectiveness of Multi-step Zooming.}
As shown in Table~\ref{tab:zoom_strategy}, performing two consecutive $\times\frac{1}{2}$ zooms consistently outperforms a single $\times\frac{1}{4}$ shrink across all domains, notably, improving overall accuracy from \textbf{62.1\%} to \textbf{63.9\%} (+1.8\%). Crucially, this self-correction arises from the model’s \textit{inherent spatial priors} rather than explicit contextual injection—a trend validated by the controlled comparisons in our supplementary tables.

\begin{table}[t]
\centering
\resizebox{\linewidth}{!}{
\begin{tabular}{lccccccc}
\toprule
\textbf{Strategy} & \textbf{Overall} & \textbf{Dev} & \textbf{Creative} & \textbf{CAD} & \textbf{Scientific} & \textbf{Office} & \textbf{OS} \\
\midrule
$1/4$ (one-step)     & 62.1 & 59.5 & 51.9 & 63.2 & 65.0 & 77.4 & 61.2 \\
\textbf{$1/2 + 1/2$ (two-step)} & \textbf{63.9} & \textbf{62.9} & \textbf{54.0} & \textbf{64.4} & \textbf{64.6} & \textbf{80.4} & \textbf{61.7} \\
\bottomrule
\end{tabular}
}
\caption{\textbf{Comparison of Zoom Strategies.}
Accuracy (\%) under two zoom configurations: Two-step $\times\frac{1}{2}$ vs. One-step $\times\frac{1}{4}$. 
The two-step strategy performs two consecutive shrink operations that each reduce the viewport to $\frac{1}{2}$ of its previous size, whereas the one-step strategy directly shrinks the viewport to $\frac{1}{4}$ of the original size in a single operation.
The two-step strategy consistently outperforms the one-step alternative across all domains.}
\label{tab:zoom_strategy}
\end{table}

\subsubsection{When to Click out} 

\begin{table*}[h]
\vspace{-5mm}
\centering
\resizebox{\textwidth}{!}{%
\small
\renewcommand{\arraystretch}{1.0}
\setlength{\tabcolsep}{3.5pt}

\begin{tabular}{l|cccc|cccc|cccc|cccc|cccc}
\toprule
\textbf{Model} &
\multicolumn{4}{c|}{\textbf{easy-normal}} &
\multicolumn{4}{c|}{\textbf{easy-mislead}} &
\multicolumn{4}{c|}{\textbf{hard-normal}} &
\multicolumn{4}{c|}{\textbf{hard-mislead}} &
\multicolumn{4}{c}{\textbf{hard-est}}\\
& d1 & d2 & d3 & d4 & d1 & d2 & d3 & d4 & d1 & d2 & d3 & d4 & d1 & d2 & d3 & d4 & d1 & d2 & d3 & d4 \\
\midrule

\rowcolor{gray!12}
\multicolumn{21}{l}{\textit{Close-sourced Models}} \\[1pt]

GPT-5 & 2.2 & 9.5 & \textcolor{red}{14.8} & \textcolor{red}{13.7} & \textcolor{red}{3.1} & \textcolor{red}{2.4} & 2.4 & 2.4 & 5.5 & 5.5 & 6.5 & 6.9 & 0.0 & \textcolor{red}{8.1} & \textcolor{red}{2.7} & 5.4 & 0.7 & 2.8 & \textcolor{red}{4.8} & \textcolor{red}{4.2} \\[2pt]

Claude-4.5-Sonnet & 30.7 & \textcolor{gray}{--} & \textcolor{gray}{--} & \textcolor{gray}{--} &
5.9 & \textcolor{gray}{--} & \textcolor{gray}{--} & \textcolor{gray}{--} &
4.9 & \textcolor{gray}{--} & \textcolor{gray}{--} & \textcolor{gray}{--} &
10.5 & \textcolor{gray}{--} & \textcolor{gray}{--} & \textcolor{gray}{--} &
6.7 & \textcolor{gray}{--} & \textcolor{gray}{--} & \textcolor{gray}{--} \\[2pt]

\midrule
\rowcolor{gray!12}
\multicolumn{21}{l}{\textit{Open-Sourced Models}} \\[1pt]

Qwen3-VL-8B & 54.2 & \textcolor{red}{89.0} & \textcolor{red}{88.5} & \textcolor{red}{87.9} & 26.2 & 38.1 & \textcolor{red}{47.6} & \textcolor{red}{42.9} & 24.9 & 55.3 & 55.3 & 56.7 & 16.2 & 32.4 & 40.5 & 45.9 & 8.1 & \textcolor{red}{19.9} & \textcolor{red}{18.8} & \textcolor{red}{18.2} \\[2pt]

Qwen3-VL-32B & 56.1 & 92.0 & \textcolor{red}{92.2} & \textcolor{red}{92.0} & 23.8 & 35.7 & 45.2 & 47.6 & 30.4 & \textcolor{red}{69.6} & \textcolor{red}{67.3} & 67.3 & 21.6 & 45.9 & 45.9 & 54.1 & 10.1 & \textcolor{red}{27.2} & \textcolor{red}{25.5} & 25.5 \\[2pt]

Qwen2.5-VL-7B & 41.1 & 61.2 & \textcolor{red}{63.5} & \textcolor{red}{63.3} & 7.1 & \textcolor{red}{9.5} & \textcolor{red}{7.1} & 9.5 & 12.9 & 21.7 & 28.1 & 28.6 & 0.0 & 10.8 & \textcolor{red}{16.2} & \textcolor{red}{13.5} & 3.1 & 8.1 & 9.0 & 9.5 \\[2pt]

UI-TARS-7B & 48.5 & 66.1 & \textcolor{red}{66.1} & \textcolor{red}{64.8} & 16.7 & 19.0 & \textcolor{red}{21.4} & \textcolor{red}{19.0} & 11.5 & 30.4 & 30.4 & 32.3 & 5.4 & 10.8 & \textcolor{red}{24.3} & \textcolor{red}{18.9} & 6.2 & 8.4 & 11.8 & 12.9 \\[2pt]

UI-TARS-1.5-7B & 62.8 & 74.0 & 77.0 & 77.3 & 19.0 & 26.2 & 28.6 & 28.6 & 21.8 & 35.0 & 37.3 & 39.2 & 18.9 & \textcolor{red}{24.3} & \textcolor{red}{21.6} & \textcolor{red}{18.9} & 6.5 & \textcolor{red}{12.0} & \textcolor{red}{11.2} & \textcolor{red}{10.4} \\[2pt]

UI-Venus-7B & 73.7 & 84.1 & \textcolor{red}{86.6} & \textcolor{red}{86.4} & 31.0 & 38.1 & 40.5 & 42.9 & 28.1 & 47.0 & 54.8 & 57.6 & 18.9 & 35.1 & \textcolor{red}{35.1} & \textcolor{red}{29.7} & 8.7 & 14.6 & 16.0 & 16.2 \\[2pt]

\bottomrule
\end{tabular}}
\vspace{-1mm}
\caption{\textbf{Comparison of Models on GUIZoom-Bench.}}
\label{tab:guizoom}
\vspace{-3mm}
\end{table*}
\begin{table}[t]
\vspace{2mm}
\centering
\resizebox{0.9\linewidth}{!}{
\begin{tabular}{lccccccc}
\toprule
\textbf{Model} & \textbf{0} & \textbf{384} & \textbf{512} & \textbf{640} & \textbf{768} & \textbf{896}  &\textbf{1024}\\
\midrule
Qwen3-VL-32B & 65.2 & 69.6 & 70.4 & 71.0 & 72.1 & 72.2&\textbf{72.3}\\
UI-Venus-7B& 65.5& \textbf{65.7} & 65.5 & 65.6 & 63.9 & 62.9  &63.3\\
\bottomrule
\end{tabular}
}
\vspace{-2mm}
\caption{\textbf{Ablations on \texttt{min\_crop\_size} for Two Models.}
Numbers are overall localization accuracy (\%). }
\label{tab:mincrop_ablation}
\vspace{-5mm}
\end{table}

\textit{The effectiveness of minimum crop size.} According to Table \ref{tab:mincrop_ablation},  two types of models exhibit markedly different sensitivities to min\_crop\_size: Qwen3-VL-32B shows a progressive accuracy increase as the minimum crop size grows, whereas UI-Venus-7B reaches its peak at smaller crop sizes and degrades consistently as min\_crop\_size increases.

\vspace{0.5em}
\noindent
\textbf{Analysis}
The contrasting trends between general-purpose VLMs and UI-specialized models stem from their different \textit{adaptability to zoom}. General VLMs benefit from larger crops because their reasoning depends heavily on global context, whereas UI-specialized models—optimized for small-window icon and widget recognition—perform best with tightly focused crops and quickly lose discriminative power as more background enters the view. This also explains the \textit{mislead} phenomenon in Table \ref{tab:guizoom}, where models that rely on localized icon cues are more easily drawn toward visually salient but irrelevant regions. These behaviors validate our benchmark taxonomy by revealing each model family’s intrinsic strengths and weaknesses under zoom. They further point to a promising direction for future work: grounding models should incorporate \textbf{multi-resolution or multi-scale training} to better handle zoom-induced context changes and make the zoom pipeline itself more robust.

\subsection{GUIZoom-Bench Results}
\textit{(1) Validation of our benchmark construction.} Across model families, the curves consistently reveal the core challenges of zooming: most models surge in early depths (e.g., Qwen3-VL-32B improves from \textbf{56.1\%→92.0\%} on \textit{easy-normal}) yet lose stability once the window becomes too narrow. In contrast, UI-specialized models—though strong on icon recognition—remain highly vulnerable to distractors, as reflected by their steady drop on \textit{hard-mislead} (e.g., UI-Venus-7B falls from \textbf{35.1\%→29.7\%} across d1→d4). These patterns confirm that our taxonomy captures both the shared pitfalls of zoom-based grounding and the model-specific weaknesses.

\vspace{2mm}
\noindent
\textit{(2) Differences in upper-bound potential across models.} The varying gain profiles across depths also reveal clear differences in model capacity. General-purpose VLMs profit more from iterative narrowing (e.g., Qwen3-VL-32B gains \textbf{+36.0\%} on \textit{hard-normal} from d1→d2), whereas UI-centric models plateau earlier with much smaller deltas, reflecting strong local sensitivity but weaker global reasoning. These disparities indicate substantial headroom for model design depending on whether robustness or precision is the priority

\vspace{2mm}
\noindent
\textit{(3) Implications for future training strategies.} The results suggest that next-generation grounding models should adopt \textit{zoom-aware training}—\eg multi-resolution supervision or dynamic cropping—to better handle scale changes. Improving robustness across context scales is essential for raising the performance ceiling of zoom-and-click pipelines and reducing mislead-related failures.

\section{Conclusion}
In this work, we revisit \textbf{\textit{zoom}} as a structured mechanism that shapes how models perceive high-resolution GUIs, rather than auxiliary post-processing tools. We introduce \method{}, a simple yet practical zoom paradigm that leverages a model’s inherent spatial priors to progressively reveal fine-grained structure without relying on additional heuristics. Complementing this, GUIZoom-Bench provides a behavior-based evaluation framework that characterizes how different models respond to zoom—highlighting where it helps, where it destabilizes predictions, and how these patterns relate to difficulty and reliability. Together, \method{} and GUIZoom-Bench offer both a practical recipe for applying zoom and a diagnostic lens for understanding the dynamics behind zoom-enabled grounding.

\vspace{4mm}
\noindent
\textbf{Limitations}
{Our method is dictated by the model’s own spatial and semantic priors, meaning our performance ceiling is limited by the model’s built-in strengths and weaknesses.
{GUIZoom-Bench captures only desktop-scale conditions, that does not directly generalize to mobile interfaces or multi-step agent interaction workflows.

\vspace{4mm}
\noindent
\textbf{Social Impacts}
Our GUI-level interactions may expose sensitive on-screen information and introduce privacy risks. In addition, our approach relies on large models whose computation contributes to non-negligible energy consumption.

{
    \small
    \bibliographystyle{ieeenat_fullname}
    \bibliography{main}
}

\clearpage
\setcounter{page}{1}
\maketitlesupplementary

\subsection{Implementation Details}
We adopt a fixed zoom depth of \textbf{$T{=}3$} iterations with a shrink ratio of \textbf{$\rho{=}0.5$}, enforcing a minimum crop size of \textbf{$m{=}768$} pixels to preserve essential context. The initial pre-zoom step employs a non-overlapping \textbf{$2{\times}2$} patch grid to establish a reliable starting region. For results in main experiment, we use the \texttt{clip} boundary mode to maintain spatial alignment of localized regions. Unless specified, we use the \texttt{shift} boundary mode in ablations to ensure a fair comparison across models. 

\vspace{-3.5mm}
\paragraph{Center handling during zoom.}
When the zoom window centered at the predicted click would cross image
boundaries, we control its behavior via a \texttt{center\_mode} flag in the
cropping function. Concretely, \texttt{shift} (default) keeps the target window
size fixed and translates the window back into the valid image region whenever
it overflows. \texttt{clip} keeps the predicted center fixed and simply
intersects the window with the image bounds, so the effective crop may be
smaller but the center is unchanged. \texttt{shrink} also fixes the center, but
adaptively reduces the window width/height to the largest size that fits inside
the image, which can lead to very small fields of view for points near the
image borders.

\vspace{-3.5mm}
\paragraph{Ablations on Pre-Zoom Strategy.}
For ablations in \ref{subsubsec:prezoom}, we implement a \textit{VLM-based} variant: we replace the heuristic distance selection with an explicit vision-language reasoning procedure. Specifically, we first invoke the VLM to perform a global grounding prediction on the original screenshot, producing one candidate location. Next, the same screenshot is evenly partitioned into a $2 \times 2$ grid, and the VLM is independently queried on each of the four patches to obtain four additional local predictions. This results in five candidate points in total. All five candidates are then projected back onto the original screenshot and visualized using identical circular markers (red filled circles) with uniform size, each annotated with a unique numerical index. A final VLM query is issued over this annotated image, prompting Qwen3-VL-32B to output the index corresponding to the point it considers most likely to represent the correct target location for the first-round grounding. The selected point is then used as the pre-zoom reference for subsequent cropping. Our default \textit{distance-based} pre-zoom uses the same five candidates but simply chooses the patch-center nearest to the global prediction.

\vspace{-2mm}
\paragraph{Ablations on Context Injection.}
For ablations on context information, we select UI-TARS-1.5-7B as our base model. We design five default context-retention settings, all sharing the same zoom and cropping strategy, but differing only in how information from the previous round is conveyed to the current round:
\begin{enumerate}
    \item \textit{Prompt Only (Positive).} The relative position of the previous click within the current cropped region is explicitly described in the prompt, accompanied by positive guidance encouraging the model to refine and correct its previous prediction.
    \item \textit{Prompt Only (Neutral).} The relative position of the previous click is provided in the prompt as plain information, without any evaluative or guiding language, merely indicating that it corresponds to the last-round result.
    \item \textit{Visual Marking.} The previous click is annotated directly on the current image using a red cross marker, serving as a visual cue without textual explanation.
    \item \textit{Full Context Injection.} Both the previous round’s input image and its corresponding click point are supplied as auxiliary context alongside the current image, enabling the model to reason jointly over cross-round visual information.
    \item \textit{No Context.} No information from the previous round is provided; each round is performed independently.
\end{enumerate}

\vspace{3mm}
\subsection{More Ablations}
\paragraph{The Effectiveness of Context.}
Table~\ref{tab:context_ablation} shows that the highest overall accuracy is achieved when no contextual information is injected across zoom rounds (54.1\%), while all context-aware variants lead to performance degradation, most notably the Full Context Injection setting (19.0\%). Among them, Visual Marking (Red Cross) exhibits the least drop, indicating that purely visual cues introduce less disturbance than prompt-based or multi-modal context, whereas textual descriptions of previous predictions consistently bias the model away from optimal visual reasoning. These results highlight a core limitation of training-free zoom-based grounding: externally introduced context shifts the model’s implicit spatial distribution and often anchors it to earlier errors, leading to error accumulation during iterative refinement. Crucially, this does not suggest that context is inherently harmful, but rather that the model lacks the learned capacity to interpret such information when injected only at inference time. This motivates \textit{treating context as a trainable guidance signal~\cite{Zu2024ept, li2023visualincontextprompting} instead of an ad-hoc heuristic}, enabling future models to explicitly learn how to leverage historical localization traces as structured auxiliary information to improve robustness and iterative correction.

\begin{table}[t]
\centering
\small
\setlength{\tabcolsep}{6pt}
\resizebox{\columnwidth}{!}{
\begin{tabular}{l c c c c c c c}
\toprule
\textbf{Context Strategy} & \textbf{Overall} & DEV & CREATIVE & CAD & SCIENTIFIC & OFFICE & OS \\
\midrule
None (No Context) & \textbf{54.1} & 48.8 & 50.4 & 42.1 & 57.1 & 77.0 & 54.1 \\
Prompt Only (Neutral) & 49.8 & 47.8 & 44.6 & 37.5 & 53.5 & 72.6 & 46.9 \\
Prompt Only (Positive) & 48.3 & 47.8 & 44.9 & 31.4 & 51.2 & 71.3 & 46.9 \\
Visual Marking (Red Cross) & 51.0 & 49.2 & 45.2 & 38.7 & 53.5 & 73.0 & 51.0 \\
Full Context Injection & 19.0 & 18.1 & 20.2 & 18.0 & 20.5 & 22.6 & 13.3 \\
\bottomrule
\end{tabular}}
\caption{\textbf{Effect of Context Retention Strategies.} 
Each row corresponds to one of the five designed context mechanisms.}
\label{tab:context_ablation}
\end{table}

\paragraph{The Effectiveness of Boundary Handling Mode.}
Across all three models in Table~\ref{tab:center_mode}, shrink consistently yields the lowest accuracy (e.g., 63.6\% vs.\ 72.1\% for Qwen3-VL-32B), while shift and clip remain close. This is consistent with their geometric behavior: both shift and clip preserve a reasonably large field of view—either by translating the fixed-size crop back into the image (shift) or by intersecting it with the image bounds (clip). In contrast, shrink keeps the predicted center fixed and adaptively reduces the crop to the largest window that fits within the image. For points near the image borders (which are common in GUI layouts, e.g., toolbars and sidebars), this produces very small crops that discard most contextual information. As a result, subsequent iterations are forced to refine within an overly narrow and potentially misaligned region, amplifying early localization errors instead of correcting them. This systematic loss of context explains why shrink is consistently the worst-performing strategy.

\begin{table}[t]
\centering
\small
\setlength{\tabcolsep}{6pt}
\resizebox{0.8\columnwidth}{!}{
\begin{tabular}{lccc}
\toprule
\textbf{Model} & \textbf{Shift} & \textbf{Clip} & \textbf{Shrink} \\
\midrule
Qwen3-VL-32B          & 72.1 & 69.6 & 63.6 \\
UI-Venus-72B          & 73.0 & 72.8 & 70.4 \\
UI-Venus-72B (thres=50) & 72.9 & 73.1 & 71.2 \\
\bottomrule
\end{tabular}}
\caption{\textbf{Effect of different Center-Handling Strategies.}}
\vspace{-4mm}
\label{tab:center_mode}
\end{table}

\paragraph{The Selection of Distance Threshold.}
The distance reported here denotes the Euclidean pixel displacement between two consecutive predicted click locations produced in adjacent zoom rounds, measured in the original image space. Consecutive pairs are categorized as 1–1 (correct pairs) when both predictions fall inside the ground-truth bounding box, and as 0–1 / 1–0 (error pairs) when only one of them is correct, indicating unstable refinement. As shown in the table, correct pairs exhibit tightly clustered distances near zero (median $<$ 10 px), reflecting smooth local refinement, whereas error pairs show substantially larger displacements (median often $>$ 50 px) with heavy tails, corresponding to abrupt spatial jumps. The optimal thresholds across individual stages lie within a narrow range (33.4–52.2 px), and the aggregated distribution yields an optimal threshold of 50.7 px with 91.8\% accuracy, motivating our adoption of 50 px as a unified decision boundary. This threshold effectively separates stable, convergent refinement behavior from erroneous drift, providing a simple yet robust criterion for identifying trustworthy iterative localization without additional supervision.

\begin{table}[t]
\centering
\small
\setlength{\tabcolsep}{6pt}
\resizebox{\columnwidth}{!}{
\begin{tabular}{lcccc}
\toprule
\textbf{Pair} & \textbf{Best Threshold (px)} & \textbf{Accuracy} & \textbf{Correct Mean (px)} & \textbf{Error Mean (px)} \\
\midrule
1--2 & 33.4 & 88.2\% & 13.7 & 284.9 \\
2--3 & 47.8 & 91.1\% & 8.9  & 97.8 \\
3--4 & 52.2 & 97.0\% & 3.5  & 104.9 \\
\midrule
All (1--4) & \textbf{50.7} & \textbf{91.8\%} & 8.2 & 207.0 \\
\bottomrule
\end{tabular}}
\caption{\textbf{Optimal distance threshold for correctness classification between consecutive predictions.}
We report the best pixel threshold that separates correct and error pairs, along with the resulting accuracy.}
\vspace{-2mm}
\label{tab:threshold_selection}
\end{table}

\subsection{Performance Between Zoom-Based Methods}
\begin{table*}[t]
\centering
\small
\setlength{\tabcolsep}{5.5pt}
\begin{tabular}{l c ccccccc}
\toprule
\textbf{Method w/ Qwen2.5-VL-7B} & \textbf{Training-Free} & \textbf{Dev} & \textbf{Creative} & \textbf{CAD} & \textbf{Scientific} & \textbf{Office} & \textbf{OS} & \textbf{Ovr} \\
\midrule
GUI-Spotlight & \xmark & 29.8 & 29.1 & 39.2 & 39.8 & 63.9 & 24.5 & 38.7 \\
GUI-Cursor    & \xmark & 68.9 & 42.7 & 46.7 & 61.4 & 74.8 & 50.0 & 56.5 \\
ReGUIDE       & \xmark & --   & --   & --   & --   & --   & --   & 44.4 \\
RegionFocus   & \cmark & 29.1 & 27.0 & 22.2 & 37.0 & 51.7 & 29.6 & 32.1 \\
Ours          & \cmark & 42.1 & 39.6 & 32.2 & 43.7 & 61.3 & 50.0 & 44.0 \\
\bottomrule
\end{tabular}
\caption{\textbf{Comparison of zoom-based grounding methods with Qwen2.5-VL-7B.}
\cmark\ indicates methods that require task-specific training, while \xmark\ denotes training-free approaches.}
\label{tab:qwen25_comparison}
\end{table*}

As shown in Table~\ref{tab:qwen25_comparison}, our \textit{training-free} method achieves an overall accuracy of \textbf{44.0\%} with \textit{Qwen-2.5-VL-7B}, substantially outperforming the previous training-free method RegionFocus (\(32.1\%\)) and the finetuned method GUI-Spotlight(\(38.7\%\)), and closely matching ReGUIDE, which relies on additional training. Notably, while GUI-Cursor benefit from explicit training and reach higher performance, our approach demonstrates that carefully designed zoom strategies alone can already recover a large portion of the performance gap without any parameter updates.

\vspace{2mm}
\noindent
\textbf{Insight.} These results indicate that the performance gains primarily originate from better utilization of the zoom mechanism itself, suggesting that the zoom process still contains untapped potential even under purely inference-time settings. At the same time, the performance margin between GUI-Cursor and all training-free methods implies that further improvements are likely to be achieved when zoom strategies are explicitly optimized during training. This naturally points to reinforcement learning or other policy optimization paradigms as promising directions for learning more effective and adaptive zoom behaviors, enabling the model to go beyond heuristic refinement and approach the upper bound demonstrated by trained counterparts.

\subsection{Prompts}

\begin{tcolorbox}[
  colback=gray!5,
  colframe=black!40,
  title={Prompt for Candidate Selection in Pre-zoom},
  fonttitle=\bfseries
]
\ttfamily
You are given a UI screenshot and a user command: \{instruction\}. \\
There are \{num\_candidates\} candidate UI elements: \{candidates\}. \\
Based on the command and the description of candidates, choose which candidate better matches the command. \\
Output only the most preferred coordinate in the format [x1, x2, y1, y2].
\end{tcolorbox}

\vspace{0.5mm}
\begin{tcolorbox}[
  colback=gray!5,
  colframe=black!40,
  title={Prompt for UI-Venus-7B Inference},
  fonttitle=\bfseries
]
\ttfamily
Outline the position corresponding to the instruction: \{instruction\}. \\
The output should be only [x1, y1, x2, y2].
\end{tcolorbox}

\vspace{-2mm}
\begin{tcolorbox}[
  colback=gray!5,
  colframe=black!40,
  title={Tool-call Prompt for Qwen3-VL Grounding},
  fonttitle=\bfseries
]
\ttfamily
System: \\
You are a helpful assistant that can click on elements in UI screenshots. \\
You are provided with a set of available tools inside XML tags: \\
\texttt{<tools>\{...\}</tools>}. Each tool is specified by a JSON function signature. \\
IMPORTANT: You \textbf{must} respond by calling one of these tools to click on the requested element. \\
For each function call, you must return a JSON object wrapped in \texttt{<tool\_call>...</tool\_call>} tags: \\
\texttt{<tool\_call>} \\
\texttt{\{"name": <function-name>, "arguments": <args-json-object>\}} \\
\texttt{</tool\_call>} \\
Example: To left-click at coordinates (500, 300), respond with: \\
\texttt{<tool\_call>} \\
\texttt{\{"name": "computer\_use", "arguments": \{"action": "left\_click", "coordinate": [500, 300]\}\}} \\
\texttt{</tool\_call>} \\[4pt]

User: \\
(1) A UI screenshot image. \\
(2) An instruction: \\
\texttt{"Please click on the element described as: \{instruction\}.} \\
\texttt{Respond with a tool call containing the exact pixel coordinates."}
\end{tcolorbox}

\subsection{Pseudocode of Our Method}
\vspace{-0.5mm}
\begin{algorithm}[]
\small
\caption{ZoomClick: Patch–Global Pre-zoom with Drift-Free Narrowing}
\label{alg:zoomclick}
\KwIn{image $I$ ($W{\times}H$), instruction $q$}
\KwOut{final click $p_{\text{px}}$}
\textbf{Params:} $K{=}2{\times}2$, threshold $\tau$ (px), shrink $\rho\!\in\!(0,1)$, min-crop $m$, max steps $T$, mode $\in\{\texttt{shift},\texttt{clip},\texttt{shrink}\}$\;

$V\!\leftarrow\!(0,0,1,1)$,\quad $I_1\!\leftarrow\!I$,\quad $(W_1,H_1)\!\leftarrow\!(W,H)$\;

\BlankLine
\textbf{Pre-zoom (patch–global consensus)}\;
\tcc{Core-1: Selects a cleaner local context when patch and global predictions agree.}
$p_{\text{dir}}\!\leftarrow\!\mathcal{G}(I_1,q)$;\quad generate non-overlapping patches $\{I^{(k)}\}_{k=1}^{K}$\;
\For{$k=1$ \KwTo $K$}{
  $\hat p^{(k)}\!\leftarrow\!\mathcal{G}(I^{(k)},q)$\;
  $p^{(k)}\!\leftarrow\!\text{MapToOrig}(\hat p^{(k)},\text{ patch }k)$\;
  $d_k\!\leftarrow\!\|p_{\text{dir}}-p^{(k)}\|_2$;
}
$k^\star\!\leftarrow\!\arg\min_k d_k$;\quad
$p^{(1)}\!\leftarrow\!\begin{cases}
p^{(k^\star)}, & d_{k^\star}\!<\!\tau\\
p_{\text{dir}}, & \text{otherwise}
\end{cases}$\;

\BlankLine
\textbf{Iterative narrowing (min-crop context retention)}\;
\tcc{Core-2: Multi-step zoom iteratively narrows the view to refine localization and reach the precise target.}
\For{$t=1$ \KwTo $T$}{
  $p_r^{(t)}\!\leftarrow\!\text{MapToOrig}(p^{(t)},V)$\;
  \If{$t{=}T$}{\Return $p_{\text{px}}\!\leftarrow\!\text{ToPixels}(p_r^{(t)})$}
  $(\tilde W,\tilde H)\!\leftarrow\!(\max(\lfloor \rho W_t\rfloor,m),\max(\lfloor \rho H_t\rfloor,m))$\;
  $(I_{t+1},B_t)\!\leftarrow\!\text{CropFromOrig}(I,\,p_r^{(t)},\,\tilde W,\,\tilde H,\,\text{mode})$\;
  \tcc{Core-3: Min-crop preserves contextual cues; boundary mode controls zoom behavior.}
  $V\!\leftarrow\!\text{ComposeViewport}(V,B_t)$;\quad $(W_{t+1},H_{t+1})\!\leftarrow\!\text{Size}(I_{t+1})$\;
  $p^{(t+1)}\!\leftarrow\!\mathcal{G}(I_{t+1},q)$;
}
\end{algorithm}

\onecolumn
\subsection{More Visualization}
\vspace{3mm}
While the main paper demonstrates the effectiveness of ZoomClick in correcting previously mislocalized predictions through progressive zooming, we further present several misled cases to illustrate its current limitations. As visualized, two representative failure patterns are observed. First, although a minimum crop size is enforced and the relevant contextual region remains preserved within the view, the model may \textit{fail to exploit this context when the cropped distribution deviates from its familiar training distribution}, leading to ineffective contextual reasoning and incorrect localization. Second, our method does not explicitly process or restructure the input instruction. For instructions involving sequential or relative semantics (e.g., first, oldest, or spatial comparison among adjacent elements*), the model \textit{struggles to correctly interpret the referential intent and is prone to being misled by visually similar neighboring targets}. These cases reveal that beyond visual refinement, improved linguistic understanding~\cite{ma2025surveyimagequalityassessment, xie2024unifyingunderstandinggenerationera} and distribution-aware adaptation remain key directions for future enhancement.

\vspace{5mm}
\begin{figure*}[h]
    \centering
    \includegraphics[width=\linewidth]{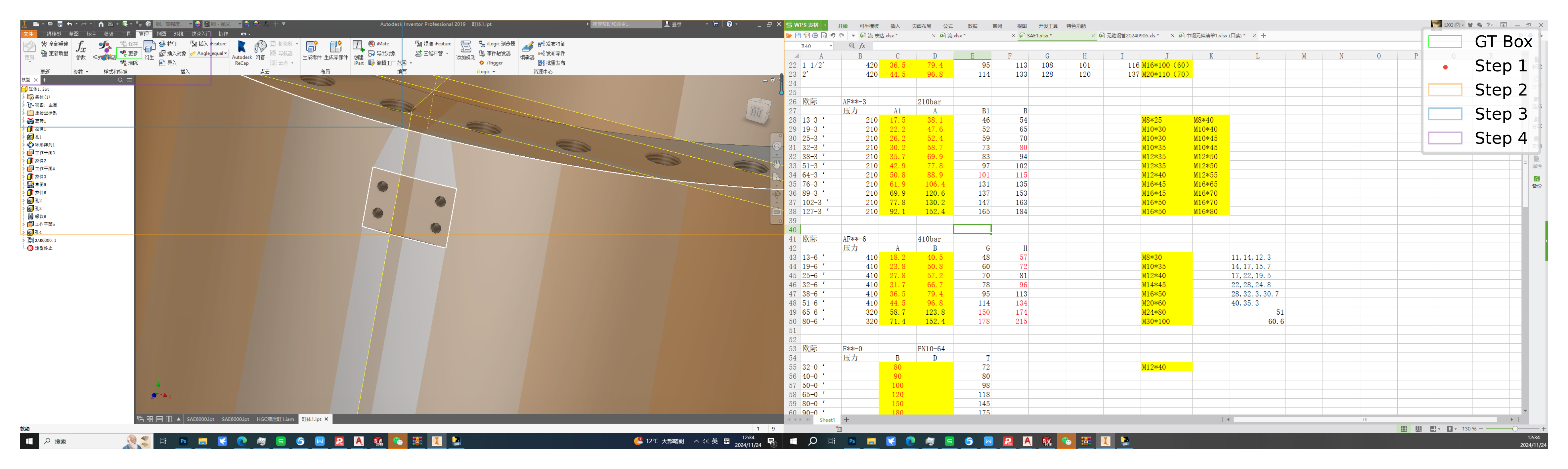}
    \caption{\textit{"Update style."}}
    \label{fig:vis3}
\end{figure*}

\begin{figure*}[h]
    \centering
    \includegraphics[width=\linewidth]{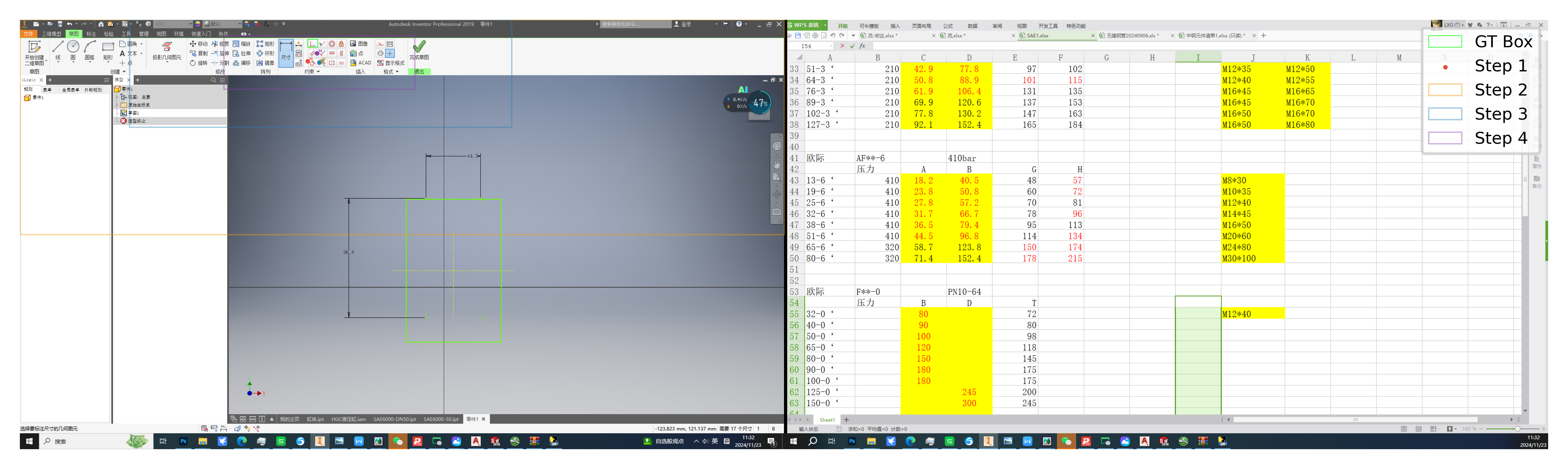}
    \caption{\textit{"Overlap Constraint."}}
    \label{fig:vis2}
\end{figure*}

\paragraph{Case 1: Context preserved but ineffective due to distribution shift.}
Figure \ref{fig:vis3} and \ref{fig:vis2} represent cases where the minimum crop size successfully retains sufficient surrounding context, and the target region remains fully visible within the view. However, the model still produces incorrect localization. Visually, these cases exhibit atypical visual layouts or uncommon density patterns compared to the model’s training distribution, such as irregular spatial arrangements, cluttered local regions, or uncommon object co-occurrence. Despite the presence of informative context, the model fails to leverage it for disambiguation, resulting in predictions that drift toward visually salient but semantically irrelevant regions. This suggests that context preservation alone is insufficient when the cropped view deviates from the learned distribution manifold~\cite{zu2025rift}.

\begin{figure*}[h]
    \centering
    \includegraphics[width=\linewidth]{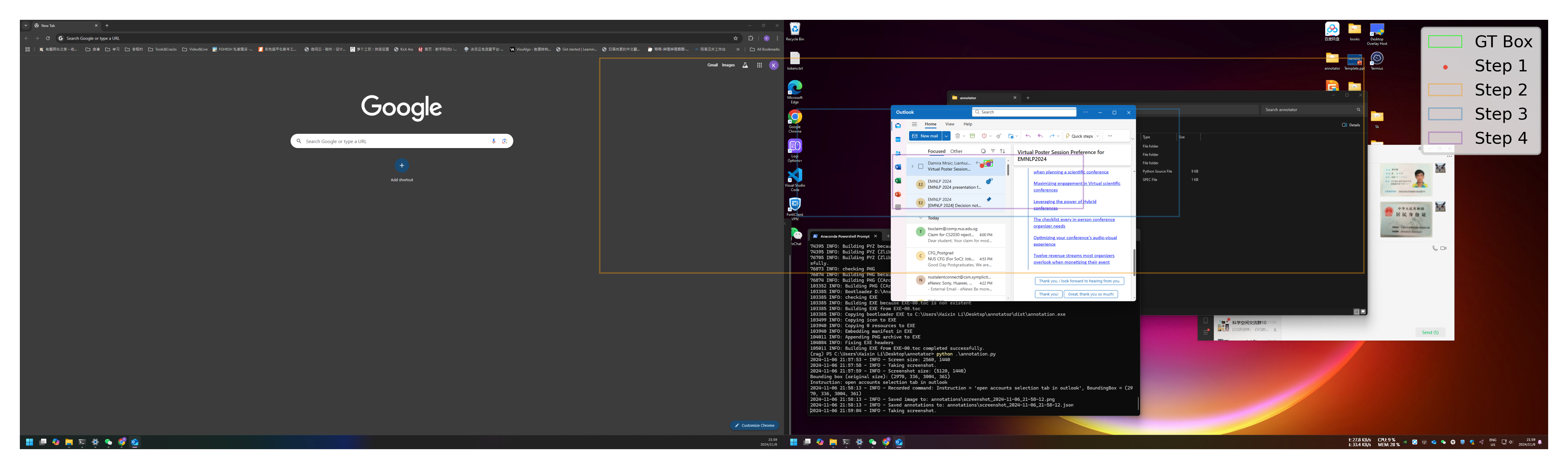}
    \caption{\textit{"Unpin the first mail entry."}}
    \label{fig:vis1}
\end{figure*}

\begin{figure*}[h]
    \centering
    \includegraphics[width=\linewidth]{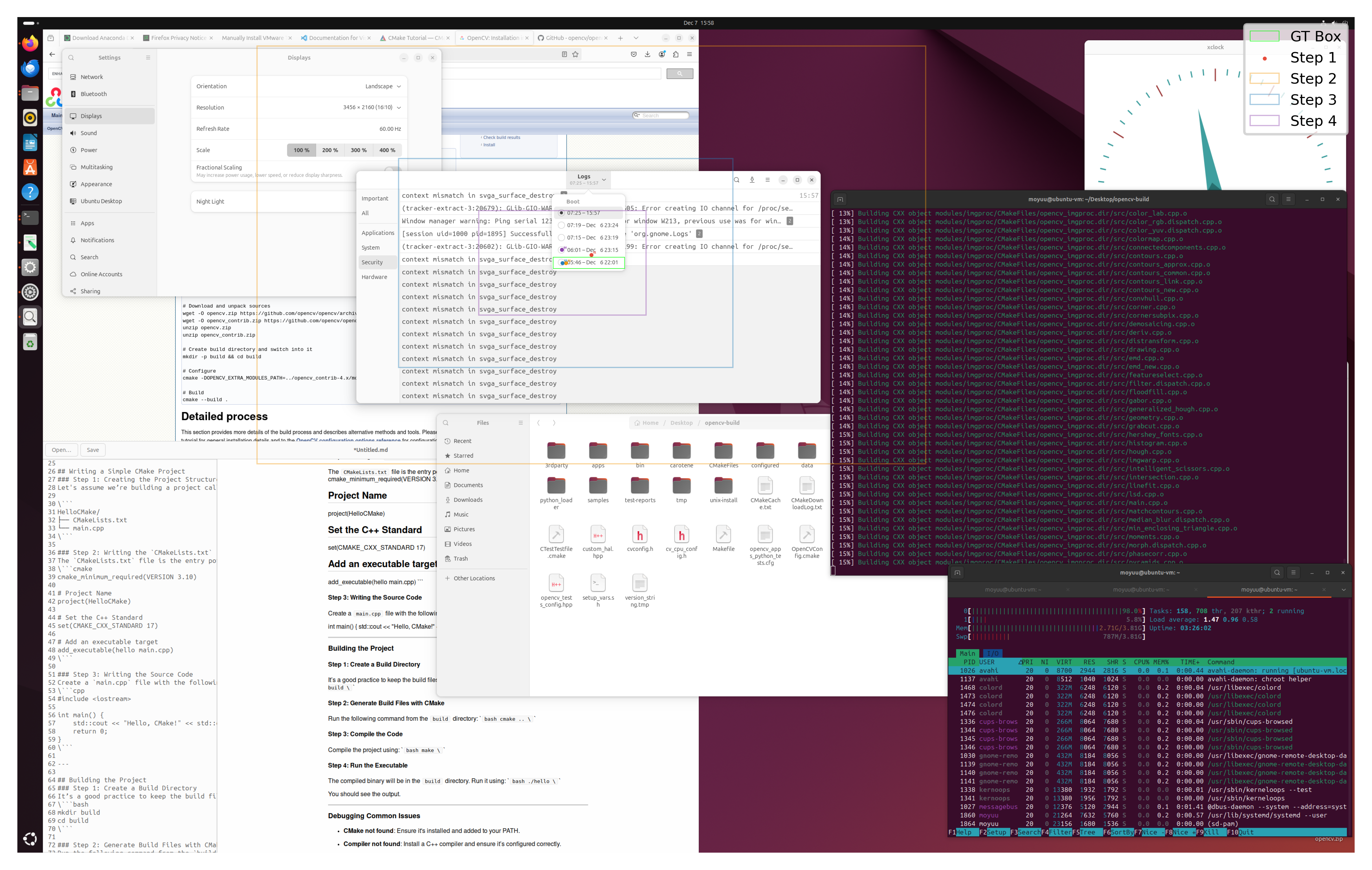}
    \caption{\textit{"View oldest logs."}}
    \label{fig:vis4}
\end{figure*}

\paragraph{Case 2: Ambiguous instruction with sequential or comparative semantics.}
Figure \ref{fig:vis1} and \ref{fig:vis4} focus on instruction-driven failures involving sequential or relative semantics, such as “first”, “last”, or references based on positional comparison. In these cases, the visualization shows multiple visually similar elements clustered in close proximity, where the true target differs mainly by ordering or relational position rather than appearance. The model’s prediction tends to gravitate toward the most visually prominent or centrally located candidate, instead of the semantically correct one indicated by the instruction. This reveals a limitation in understanding instruction-dependent relational cues, causing the system to be misled by nearby distractors with strong visual similarity.


\end{document}